\pdfoutput=1 
\documentclass[final]{cvpr}

\usepackage{times}
\usepackage{epsfig}
\usepackage{graphicx}
\usepackage{amsmath}
\usepackage{amssymb}
\usepackage{capt-of}
\usepackage{float}

\usepackage[pagebackref=true,breaklinks=true,colorlinks,bookmarks=false]{hyperref}

\def\cvprPaperID{****} 
\def\confYear{CVPR 2021}

\begin{document}
\title{Dressing in Order: Recurrent Person Image Generation for Pose Transfer, Virtual Try-on and Outfit Editing}

\author{Aiyu Cui \qquad Daniel McKee \qquad  Svetlana Lazebnik \\
University of Illinois at Urbana-Champaign \\
{\tt\small \{aiyucui2,dbmckee2,slazebni\}@illinois.edu } \\ 
 \tt \url{https://cuiaiyu.github.io/dressing-in-order}
}

\maketitle



\begin{abstract}
   We propose a flexible person generation framework called Dressing in Order (DiOr), which supports 2D pose transfer, virtual try-on, and several fashion editing tasks. The key to DiOr is a novel recurrent generation pipeline to sequentially put garments on a person, so that trying on the same garments in different orders will result in different looks. Our system can produce dressing effects not achievable by existing work, including different interactions of garments (e.g., wearing a top tucked into the bottom or over it), as well as layering of multiple garments of the same type (e.g., jacket over shirt over t-shirt). DiOr explicitly encodes the shape and texture of each garment, enabling these elements to be edited separately. Joint training on pose transfer and inpainting helps with detail preservation and coherence of generated garments. Extensive evaluations show that DiOr outperforms other recent methods like ADGAN \cite{adgan} in terms of output quality, and handles a wide range of editing functions for which there is no direct supervision. 
\end{abstract}

\section{Introduction}
Driven by the power of deep generative models and commercial possibilities, person generation research has been growing fast in recent years. Popular applications include virtual try-on \cite{mgvton,viton,cagan,vogue,amazon-unpaired,cpvton,yang2020towards}, fashion editing \cite{dong2020fashion,hsiao2019fashion++}, and pose-guided person generation \cite{esser2018variational,han2019clothflow,li2019dense, liu2019liquid,ma2017pose, gfla,sarkar2021style,nhrr,siarohin2018deformable,tang2020xinggan, patn}. 
Most existing work addresses only one generation task at a time, despite similarities in overall system designs. 
Although some systems \cite{han2019clothflow, adgan, sarkar2021style, nhrr} have been applied to both pose-guided generation and virtual try-on, they lack the ability to preserve details \cite{adgan,sarkar2021style} or lack flexible representations of shape and texture that can be exploited for diverse editing tasks \cite{han2019clothflow,adgan,sarkar2021style,nhrr}.

\begin{figure}
    \includegraphics[width=1\linewidth]{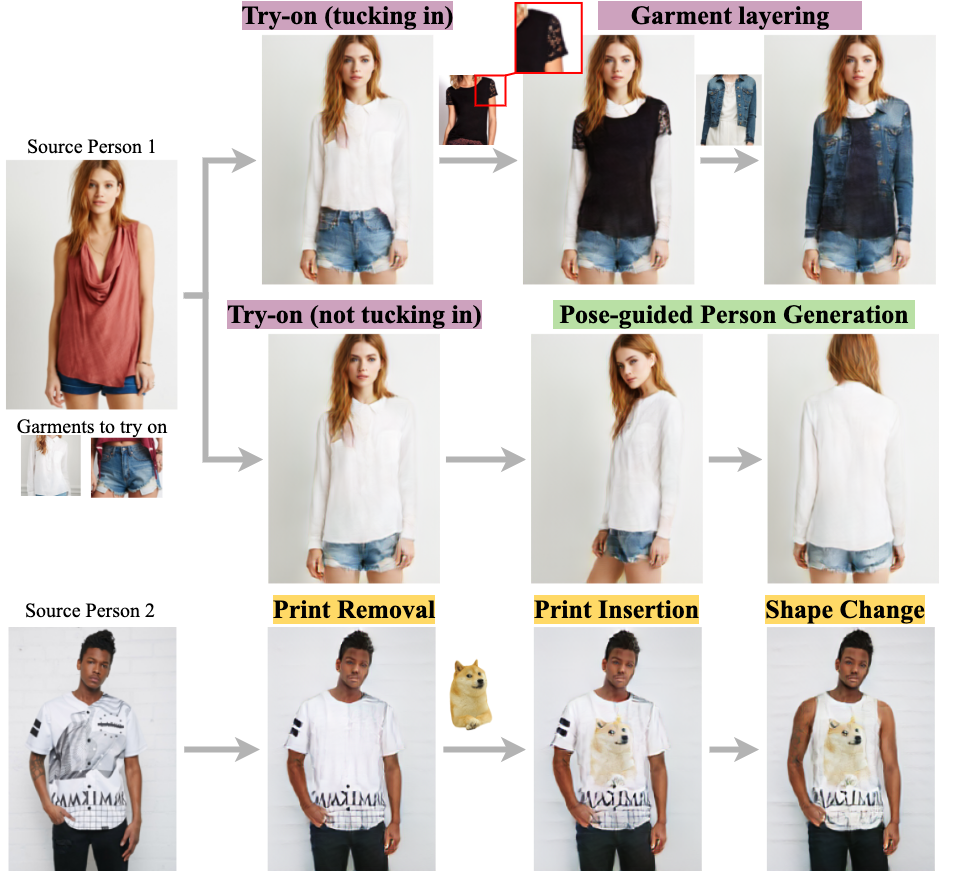}
    \label{fig:front}
    \vspace*{-15pt}
    \caption{Applications supported by our DiOr system: Virtual try-on supporting different garment interactions (tucking in or not) and overlay; pose-guided person generation; and fashion editing (texture insertion and removal, shape change). Note that the arrows indicate possible editing sequences and relationships between images, {\em not} the flow of our system. }
    \vspace*{-15pt}
    \centering 
\end{figure}

\begin{figure*}[ht]

    \vspace*{-5pt}
    \centering
    \includegraphics[width=0.92\linewidth]{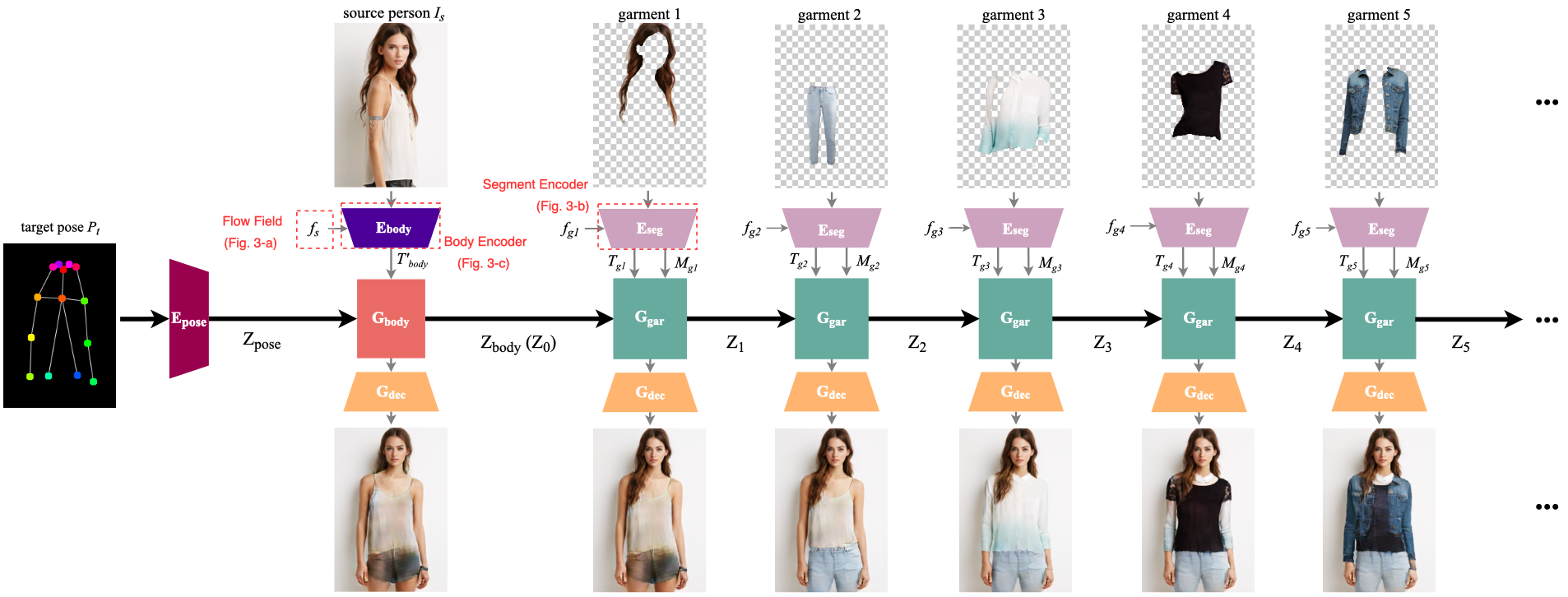}
    \caption{\textbf{DiOr generation pipeline} (see Section 3 for details). We represent a person as a (\textit{pose}, \textit{body}, \textit{\{garments\}}) tuple. Generation starts by encoding the target pose as $Z_{pose}$ and the source body as texture map $T_{body}$. Then the body is generated as $Z_{body}$ by the generator module $\mathbf{G}_\mathrm{body}$. $Z_{body}$ serves as $Z_0$ for the recurrent garment generator $\mathbf{G}_\mathrm{gar}$, which receives the garments in order, each encoded by a 2D texture feature map $T_{gk}$ and soft shape mask $M_{gk}$. In addition to masked source images, the body and garment encoders take in estimated flow fields $f$ to warp the sources to the target pose. We can decode at any step to get an output showing the garments put on so far. }
    \label{fig:pipeline}
    
    \vspace*{-15pt}
\end{figure*}

This paper proposes a flexible 2D person generation pipeline applicable not only to pose transfer and virtual try-on, but also fashion editing, as shown in Figure \ref{fig:front}. The proposed pipeline is shown in Figure \ref{fig:pipeline}. We separately encode pose, skin, and garments, and the garment encodings are further separated into shape and texture. This allows us to freely play with each element to achieve different looks. 

In real life, people put on garments one by one, and can layer them in different ways (e.g., shirt tucked into pants, or worn on the outside). However, existing try-on methods start by producing a mutually exclusive garment segmentation map and then generate the whole outfit in a single step. This can only achieve one look for a given set of garments, and the interaction of garments is determined by the model. By contrast, our system incorporates a novel recurrent generation module to produce different looks depending on the order of putting on garments. This is why we call our system {\bf DiOr}, for {\bf Dressing in Order}. 

After a survey of related work in Section 2, we will describe our system in Section 3. Section 3.1 will introduce our encoding of garments into 2D shape and texture, enabling each to be edited separately. The shape is encoded using soft masks that can additionally capture transparency. A flow field estimation component at encoding time allows for a more accurate deformation of the garments to fit the target pose. Section 3.2 will describe our recurrent generation scheme that does not rely on garment labels and can handle a variable number of garments. Section 3.3 will discuss our training approach, which combines pose transfer with inpainting to enable preservation of fine details. Section 4 will present experimental results (including comparisons and user study), and Section 5 will illustrate the editing functionalities enabled by our system.

\section{Related Work}

\noindent{\bf Virtual try-on.} 
Generation of images of a given person with a desired garment on is a challenging task that requires both capturing the garment precisely and dressing it properly on the given human body. 
The simplest try-on methods are aimed at replacing a single garment with a new one \cite{mgvton, han2019clothflow,viton,cagan,vogue,kedan, cpvton, yang2020towards}. Our work is more closely related methods that attempt to model all the garments worn by a person simultaneously, allowing users to achieve multiple garment try-on \cite{li2021toward,adgan, amazon-unpaired,swapnet, nhrr}. SwapNet \cite{swapnet} works by transferring all the clothing from one person's image onto the pose of another target person. This is done by first generating a mutually exclusive segmentation mask of the desired clothing on the desired pose. 
O-VITON \cite{amazon-unpaired} also starts by producing a mutually exclusive segmentation mask for all try-on garments, and then injects the garment encodings into the associated regions. Unlike our work, O-VITON cannot change the pose of the target person. Attribute-decomposed GAN (ADGAN) \cite{adgan} encodes garments in each class into a 1D style code and feeds a concatenation of codes into a StyleGAN \cite{stylegan} generator. It additionally conditions on 2D pose, enabling pose transfer as well as try-on. Our system adopts a similar kind of conditioning. However, as will be seen in our comparative evaluation, ADGAN's 1D garment encoding, which does not separate shape from texture, is severely limited in its fidelity of garment reproduction.

Sarkar et al. \cite{sarkar2021style,nhrr} achieve high-quality try-on results by aligning the given human images with a 3D mesh model (SMPL~\cite{loper2015smpl}) via DensePose~\cite{guler2018densepose}, estimating a UV texture map corresponding to the desired garments, and rendering this texture onto the desired pose. The focus of our work is different, as we avoid explicit 3D human modeling. 

All of the above methods assume a pre-defined set of garment classes (e.g., tops, jackets, pants, skirts, etc.) and allow at most one garment in each class. This precludes the ability to layer garments from the same class (e.g., one top over another).  
By contrast, while we rely on an off-the-shelf clothing segmenter, our generation pipeline does not make use of garment classes, only masks. 
Moreover, in all previous work, when there is overlap between two garments (e.g. top and bottom), it is up to the model to decide the interaction of the two garments, (e.g., whether a top is tucked into the bottom). 
Unlike these methods, ours produce different results for different dressing orders.
\smallskip

\begin{figure}
    \centering
    \includegraphics[width=\linewidth]{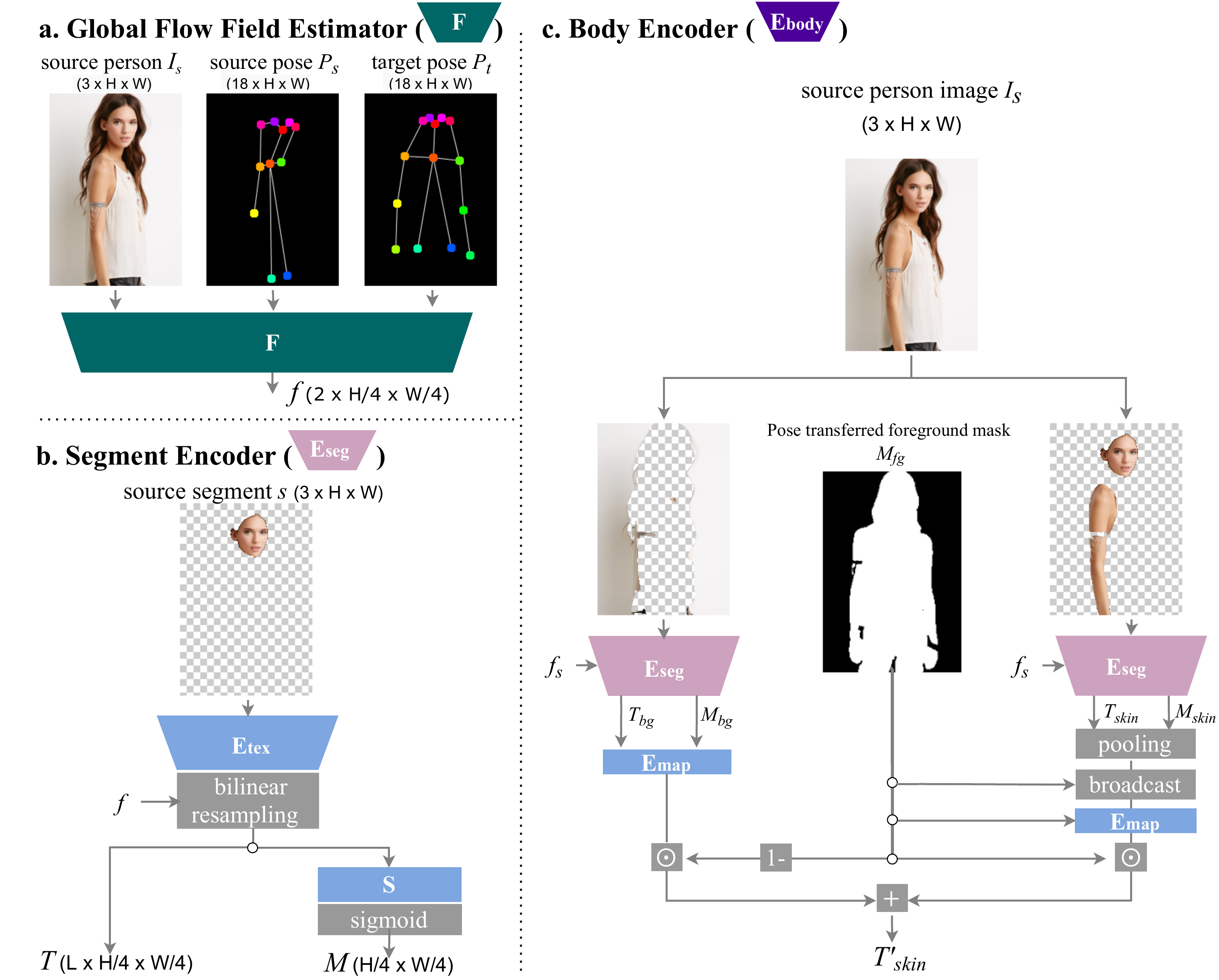}
    \caption{System details. (a) Global flow field estimator $\mathbf{F}$ adopted from GFLA \cite{gfla}, which is modified to only yield a flow field $f$. (b) Segment encoder $\mathbf{E}_\mathrm{seg}$ that produces a spatially aligned texture feature map $T$ and a soft shape mask $M$. (c) Body encoder $\mathbf{E}_\mathrm{body}$ that broadcasts a mean skin vector to the entire foreground region (union of the masks of pose-transferred foreground parts) and maps it to the correct dimension by $\mathbf{E}_\mathrm{map}$ for later style blocks. }
    \label{fig:encoder1}
\end{figure}

\noindent {\bf Pose transfer} 
requires changing the pose of a given person while keeping that person's identity and outfit the same. Several of the virtual try-on methods above \cite{han2019clothflow, adgan, swapnet, sarkar2021style, nhrr} are explicitly conditioned on pose, making them suitable for pose transfer. Our method is of this kind. An advantage of pose transfer is that there exist datasets featuring people with the same clothing in multiple poses~\cite{deepfashion}, making it easier to obtain supervision than for virtual try-on.

Most relevant to us are pose transfer methods that represent poses using 2D keypoints \cite{esser2018variational, ma2017pose, adgan, siarohin2018deformable,tang2020xinggan, patn}. However, these methods have a limited ability to capture garment details and result in blurry textures. 
 Global Flow Local Attention (GFLA) \cite{gfla} and Clothflow \cite{han2019clothflow} compute dense 2D flow fields to align source and target poses. 
 We adopt the global flow component of GFLA as part of our system, obtaining comparable results on pose transfer while adding a number of try-on and editing functions. 

Other pose transfer methods \cite{grigorev2019coordinate,li2019dense, liu2019liquid, neverova2018dense, nhrr} rely on 3D human modeling via DensePose~\cite{guler2018densepose} and SMPL \cite{loper2015smpl}. They work either by completing the UV map and re-rendering \cite{ grigorev2019coordinate, neverova2018dense, nhrr}, or learning a flow from the rich 3D information \cite{li2019dense, liu2019liquid}. Such methods represent a different philosophy from ours and are therefore less comparable.
\smallskip

\noindent {\bf Fashion editing.}
Fashion++ \cite{hsiao2019fashion++} learns to minimally edit an outfit to make it more fashionable, but there is no way for the user to control the changes.
Dong et al. \cite{dong2020fashion} edits outfits guided by user's hand sketches. 
Instead, our model allows users to edit what they want by making garment selections, and changing the order of garments in a semantic manner.

\section{Method}

This section describes our DiOr pipeline (Fig. \ref{fig:pipeline}).
 We introduce our person representation in Section 3.1, then describe our pipeline in Section 3.2, our training strategy in Section 3.3, and relationship to prior work in Section 3.4.
 \smallskip

\subsection{Person Representation}

We represent a person as a (pose, body, \{garments\}) tuple, each element of which can come from a different source image. Unlike other works (e.g., \cite{adgan, amazon-unpaired}) the number of garments can vary and garment labels are not used. This allows us to freely add, remove and switch the order of garments. 
Following prior work~\cite{adgan, gfla}, we represent pose $P$ as the 18 keypoint heatmaps defined in OpenPose \cite{openpose}. 

\noindent {\bf Garment representation.}
Given a source garment $g_k$ worn by a person in an image $I_{g_k} \in \mathbb{R}^{3\times H\times W}$, 
we first run an off-the-shelf human parser~\cite{li2020self} to obtain the masked garment segment $s_{g_k}$. We also obtain a pose estimate $P_{g_k}$ for the person in $I_{g_k}$ by OpenPose~\cite{openpose}. Because $P_{g_k}$ is different from the desired pose $P$, we need to infer a flow field $f_{g_k}$ to align the garment segment $s_{g_k}$ with $P$. We do this using the Global Flow Field Estimator $\mathbf{F}$ from GFLA \cite{gfla} (Fig. \ref{fig:encoder1}(a)).
$\mathbf{F}$ can also work on the shop images of garments only (without a person wearing them), in which case $P_{g_k}$ will just be empty heatmaps (see second example in Fig. \ref{fig:tryon}). 


Next, as shown in Fig. \ref{fig:encoder1}(b), we encode the garment segment $s_{g_k}$ by the segment encoder module $\mathbf{E}_\mathrm{seg}$. This starts with a texture encoder $\mathbf{E}_\mathrm{tex}$, which consists of the first three layers of the VGG encoder in ADGAN~\cite{adgan} (for a downsampling factor of 4) with leaky ReLU\cite{maas2013leakyrelu}. The output of $\mathbf{E}_\mathrm{tex}$ is warped by the flow field $f_{g_k}$ using bilinear interpolation, yielding a \textbf{texture feature map} denoted $T_{g_k}$. We also compute a \textbf{soft shape mask} of the garment segment as $M_{g_k}= \mathbf{S}(T_{g_k})$, where $\mathbf{S}$ is a segmenter consisting of three convolutional layers. The texture map $T_{g_k}$ and shape mask $M_{g_k}$ are both outputs of the segment encoder $\mathbf{E}_\mathrm{seg}$. 

Because the texture feature map $T_{g_k}$ will be used as style input for later style blocks, we map $T_{g_k}$ to the correct dimension of style blocks as $T'_{g_k} = \mathbf{E}_\mathrm{map}(T_{g_k} + \bar{T}_{g_k}, M_{g_k})$, where the mapping module, $\mathbf{E}_\mathrm{map}$, consists two convolutional layers and takes the stacked $T_{g_k}$ and $M_{g_k}$ as input. We found that adding $\bar{T}_{g_k}$, the mean vector of ${T}_{g_k}$, to the texture feature map benefits the hole filling, if the garment has large missing area.

\smallskip
\noindent {\bf Body representation.}
Fig. \ref{fig:encoder1}(c) shows the process of encoding the body of the source person from image $I_\mathrm{s} \in \mathbb{R}^{3\times H\times W}$. Based on the human segmenter \cite{li2020self}, we form masks corresponding to 
background $s_\mathrm{bg}$ and skin $s_\mathrm{skin}$ (the latter consisting of arms, legs and face). These are encoded by the a above-described segment encoder $\mathbf{E}_\mathrm{seg}$ to get 
$(T_\mathrm{bg}, M_\mathrm{bg})$ and $(T_\mathrm{skin}, M_\mathrm{skin})$, respectively.

To ensure that the body feature map spans the entire body region regardless of any garments that would cover it later, we compute a mean body vector $b$ of $T_\mathrm{skin}$ over the ROI defined by $M_\mathrm{skin}$. Then we broadcast $b$ to the pose-transferred foreground region $M_\mathrm{fg}$ (the union of the masks of all pose-transferred foreground parts), 
map the broadcasted feature map to the correct dimension by $\mathbf{E}_\mathrm{map}$, 
and combine with the mapped background feature $T'_\mathrm{bg}=\mathbf{E}_\mathrm{map}(T_\mathrm{bg}, M_\mathrm{bg})$ to get body texture map as 
\begin{equation} \label{eq:tbody}
    T'_\mathrm{body} =  M_\mathrm{fg} \odot \mathbf{E}_\mathrm{map}(M_\mathrm{fg} \otimes b, M_\mathrm{fg})  
    + (1 - M_\mathrm{fg}) \odot T'_\mathrm{bg}, 
\end{equation}
where $\otimes$ and $\odot$ denote broadcasting and elementwise multiplication, respectively.

\subsection{Generation Pipeline}
In the main generation pipeline (Fig.\ref{fig:pipeline}), we start by encoding the ``skeleton'' $P$, next generating the body from $T'_\mathrm{body}$ (eq. \ref{eq:tbody}), and then the garments from encoded texture and shape masks $(T'_{g_1}, M_{g_1}), ..., (T'_{g_K}, M_{g_K})$ in sequence.
\smallskip

\noindent {\bf Pose and skin generation.}
To start generation, we encode the desired pose $P$ using the pose encoder $\mathbf{E}_\mathrm{pose}$, implemented as three convolutional layers, each followed by instance normalization~\cite{ulyanov2017improved} and leaky ReLU~\cite{maas2013leakyrelu}.
This results in hidden pose map $Z_\mathrm{pose} \in \mathbb{R}^{L\times H/4 \times W/4}$, with $L$ as the latent channel size.

Next, we generate the hidden body map $Z_\mathrm{body}$ given $Z_\mathrm{pose}$ and the body texture map $T'_\mathrm{body}$ by a body generator $\mathbf{G}_\mathrm{body}$, implemented by two style blocks in ADGAN \cite{adgan}. Because our body texture map $T'_{body}$ is in 2D, ADGAN's adaptive instance normalization \cite{adain} in the style block is replaced by SPADE \cite{spade}, and we use $\mathbf{E}_\mathrm{map}$ described above to convert the style input to the desired dimensions.

\smallskip

\noindent{\bf Recurrent garment generation.}
Next, we generate the garments, treating $Z_\mathrm{body}$ as $Z_{0}$.
For the $k$-th garment, the garment generator $\mathbf{G}_\mathrm{gar}$ takes its mapped texture map $T'_{g_k}$ and soft shape mask $M_{g_k}$, together with the previous state $Z_{k-1}$, and produces the next state $Z_k$ as

\begin{equation}
    Z_k = \mathbf{\Phi}(Z_{k-1}, T'_{g_k}) \odot M_{g_k} + Z_{k-1} \odot (1-M_{g_k}) \,,
\end{equation}
where $\mathbf{\Phi}$ is a conditional generation module with the same structure as $\mathbf{G}_\mathrm{body}$ above. 
Note that the soft shape mask $M_{g_k}$ effectively controls garment opacity -- a novel feature of our representation. More details are in Appendix \ref{appendix:transparency}.

After the encoded person is finished dressing, we get the final hidden feature map $Z_{K}$ and output image $I_\mathrm{gen} = \mathbf{G}_\mathrm{dec}(Z_{K})$, where $\mathbf{G}_\mathrm{dec}$ is the decoder implemented as the same as the final decoder in ADGAN \cite{adgan}, consisting of residual blocks, upsampling and convolutional layers followed by layer normalization and ReLU. 

\subsection{Training}
Similar to ADGAN \cite{adgan}, we train our model on pose transfer:
given a person image $I_s$ in a source pose $P_s$, generate that person in a target pose $P_t$.  As long as reference images $I_t$ of the same person in the target pose are available, this is a supervised task. To perform pose transfer, we set the body image and the garment set to be those of the source person, and render them in the target pose.
There could be up to four separately encoded garments for a person to be added in order, so the recurrent generator gets ample training examples of various layering types and garment combinations.

We started by training a model solely on pose transfer, but observed
that it gives imprecise or inconsistent results for try-on and overlay (see Fig. \ref{fig:abl_joint}). To improve the realism of our model, we next experimented with training it for reconstruction as well as transfer, i.e., setting $P_t = P_s$ for a fraction of the training examples. Although this helped to preserve details and improved handling of garment overlaps, the resulting model could not complete missing regions in garments (e.g., regions covered by hair in the source). At length, we found inpainting, or recovery of a partially masked-out source image $I'_s$, to be a better supplementary training task, enabling detail preservation while filling in missing regions.
We combine the tasks by use a percentage $\alpha$ of the training data for inpainting and the rest for pose transfer. In our implementation, $\alpha=0.2$, and inpainting masks are generated by the free-form algorithm of Yu et al. \cite{yu2019free}.

To train on both pose transfer and inpainting, we use all the six loss terms from GFLA \cite{gfla}. Two of these are correctness and regularization loss for the predicted flow field, which are combined into the geometric loss $L_\mathrm{geo}$. 
Another three GFLA terms encourage consistency of generated and real target pairs: L1 loss, perceptual loss, and style loss. These are combined into the content loss $L_\mathrm{content}$. 
The final GFLA term is a GAN loss $L_{GAN}$, for which GFLA uses a single discriminator conditioned on pose, but we use two discriminators, one conditioned on the pose and the other on segmentation, as in ADGAN \cite{adgan}. 
Our discriminators have the same architecture as GFLA's.
We set the coefficients of these six loss terms by following GFLA.

In addition, to ensure that our shape masks capture the shape correctly, we use a pixel-level binary cross-entropy loss between the soft shape mask $M_g$ and its associated ``ground truth" segmentation (extracted by the parser~\cite{li2020self} from the target image) for each garment. 
This loss is denoted as $L_\mathrm{seg}$.  Our final, combined loss is thus given by
\begin{equation}
    L = L_\mathrm{content} + L_\mathrm{geo} + \lambda_{GAN}L_{GAN} + \lambda_\mathrm{seg} L_\mathrm{seg} \,,
\end{equation}
where we set $\lambda_\mathrm{seg}$ to 0.1 and $\lambda_\mathrm{GAN}$ to 1.

\subsection{Relationship to Prior Work}
Our system was most closely inspired by ADGAN \cite{adgan}. Like ADGAN, we separately encode each garment, condition the generation on 2D pose, and train on pose transfer. We also borrow the architecture of some ADGAN blocks, as explained above.
However, ADGAN encodes a garment into single 1D vector, but we encode a garment in shape and texture separately in 2D. Thus, DiOr allows shape and texture of individual garments to be edited separately, which is impossible in ADGAN. Our 2D encoding is better than ADGAN's 1D encoding at capturing complex spatial patterns, giving us superior results on virtual try-on, as shown in next section.
Besides, In ADGAN, after garments are separately encoded, all the embeddings are fused into a single vector, so the number and type of garments are fixed, and garment order is not preserved. By contrast, in our recurrent pipeline, garments are injected one at a time, and their number, type, and ordering can vary.


Our method also builds on GFLA~\cite{gfla} by adopting its global flow component and most of the loss terms. 
Our experiments will show that we achieve similar performance without GFLA's local attention component. 
Plus, GFLA can only handle pose transfer, while our model can solve a number of additional tasks.

A few previous methods have also recognized the potential of inpainting to help with human image generation, though they use it differently than we do. ACGPN \cite{yang2020towards} is a single garment try-on method that features an inpainting module to fuse the elements of the person to be rendered. In 3D-based person re-rendering literature, two recent approaches~\cite{grigorev2019coordinate,nhrr} use an inpainting loss to complete unseen regions of the UV texture map.

\section{Experiments}
\subsection{Implementation Details}
We train our model on the DeepFashion dataset \cite{deepfashion} with the same training/test split used in PATN \cite{patn} for pose transfer at $256\times176$ resolution. 
In implementation, we run Eq. 2 twice for each garment for better performance.
For the first 20k iterations, we use the same procedure as GFLA to warm up the global flow field estimator $\mathbf{F}$.
Meanwhile, we warm up the texture encoder $\mathbf{E}_\mathrm{tex}$ and final decoder $\mathbf{G}_\mathrm{dec}$ with $L_\mathrm{content}$ and $L_\mathrm{GAN}$ losses by encoding a masked input image with $\mathbf{E}_\mathrm{tex}$ and recovering the complete image using $\mathbf{G}_\mathrm{dec}$. 
Then for the next 150k iterations, we train the network with $\mathbf{F}$ frozen using Adam optimizer with learning rate $1e-4$. Finally, we unfreeze $\mathbf{F}$ and train the entire network end-to-end with learning rate $1e-5$ until the model converges. 
We train a {\bf small model} with $L$, the latent dimension of $Z$, set to 128 and a {\bf large model} with $L$=256, on one and two  TITAN Xp cards, respectively.

\subsection{Automatic Evaluation of Pose Transfer}

We run automatic evaluation on the pose transfer task, which is the only one that has reference images available. 
Table \ref{tab:pose_table} shows a comparison of our results with GFLA \cite{gfla} and ADGAN \cite{adgan}, both of which use the same 2D keypoints to represent the pose, have code and model publicly available, and use the same train/val split. 

\begin{figure}
    \centering
    \includegraphics[width=0.90\linewidth]{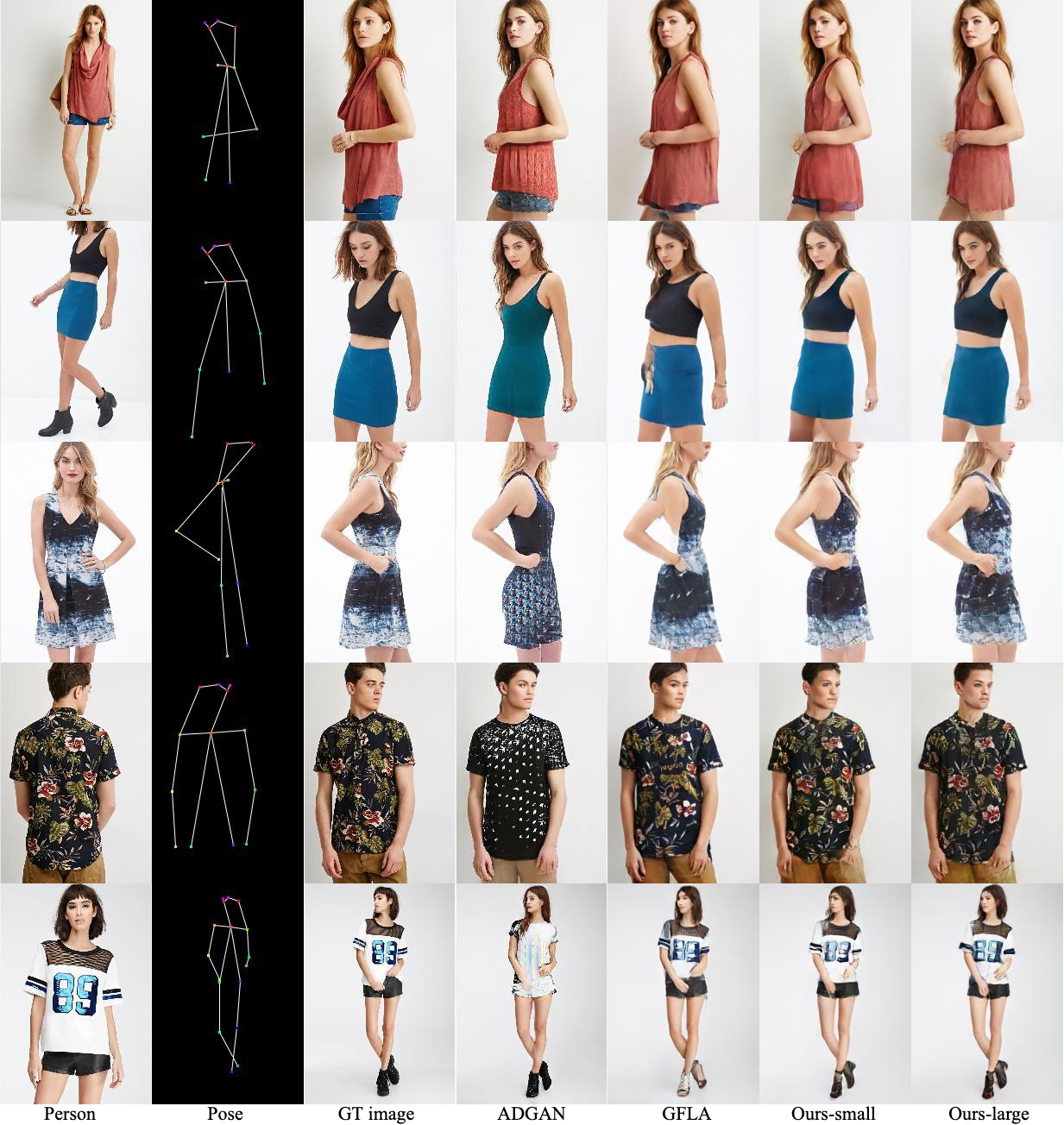}
    
    \vspace*{-5pt}
    \caption{Pose transfer results. }
    \label{fig:pose}
    \vspace*{-2pt}
\end{figure}

We compute several common metrics purporting to measure the structural, distributional, and perceptual similarity between generated and real reference images: SSIM \cite{wang2004ssim}, FID \cite{heusel2017fid}, and LPIPS \cite{zhang2018lpips}. 
Plus, we propose a new metric \textbf{sIoU}, which is the mean IoU of the segmentation masks produced by the human segmenter \cite{li2020self} for real and generated images, to measure the shape consistency. This metric is inspired by the FCN scores used in Isola et al.~\cite{isola2017pix2pix} to evaluate the consistency of label maps for the labels-to-photos generation task. To mitigate any possible bias from using the same segmenter to obtain garment masks in our pipeline, we compute this metric using ATR human parse labels \cite{liang2015atr} instead of the LIP labels \cite{liang2015lip} used in our pipeline. We further process the ATR label sets by merging left and right body parts, which gives more stable results. 
Without over-interpreting the automatic metrics (which are found to be sensitive to factors like resolution, sharpness, and compression quality of the reference images \cite{parmar2021buggy}) we can conservatively conclude that our pose transfer performance is at least comparable to that of GFLA and ADGAN, which is confirmed by the user study reported in Section 4.4. 
Our large model has the highest sIoU, which suggests its ability to preserve the structure of generated garments, and is consistent with the example outputs shown in Fig. \ref{fig:pose}. There, our output is qualitatively similar to GFLA (not surprising, since we adopt part of their flow mechanism), and consistently better than ADGAN, whose inability to reproduce garment textures or structured patterns is not obvious from the automatic metrics.
\smallskip

\begin{table}[]
\centering
\resizebox{0.95\columnwidth}{!}{%
\begin{tabular}{|l|l|l|l|l|l|}
\hline
                    & size & SSIM $\uparrow$ & FID $\downarrow$ & LPIPS $\downarrow$ & sIoU $\uparrow$  \\ 

\hline \hline
Def-GAN$^*$ \cite{siarohin2018deformable} & 82.08M &   - &  18.46 & 0.233 & - \\ 
VU-Net$^*$ \cite{esser2018variational} & 139.4M &   - &  23.67  & 0.264 & -  \\ 
Pose-Attn$^*$ \cite{patn} & 41.36M &  - &  20.74  & 0.253 & -  \\ 
Intr-Flow$^*$ \cite{li2019dense} & 49.58M  &   - &  16.31  & \textbf{0.213} & - \\ 
GFLA$^*$ \cite{gfla} & 14.04M &  0.713 &  \textbf{10.57} & 0.234 & 57.32  \\ \hline 
ours-small & 11.26M &   0.720 & 12.97 & 0.236 &  57.22  \\ 
ours-large   & 24.41M &   \textbf{0.725} & 13.10 &  0.229 & \textbf{58.63} \\ \hline
\multicolumn{4}{c}{(a) Comparisons at 256 $\times$ 256 resolution} 
\end{tabular}
}

\resizebox{0.95\columnwidth}{!}{%

\begin{tabular}{|l|l|l|l|l|l|}

\hline
& size & SSIM$\uparrow$ & FID$\downarrow$ & LPIPS$\downarrow$ & sIoU$\uparrow$  \\ \hline \hline
ADGAN \cite{adgan} & 32.29M  &   0.772 & 18.63 & 0.226 &  56.54 \\ \hline
ours-small & 11.26M  &  0.804  &  14.34 & 0.182 &  58.99 \\ 
ours-large & 24.41M &  \textbf{0.806} & \textbf{13.59} & \textbf{0.176} &  \textbf{59.99}  \\ \hline
\multicolumn{4}{c}{(b) Comparisons at 256$\times$176 resolution} 
\end{tabular}
}
\caption{
Pose transfer evaluation.  
(a) Comparison with GFLA \cite{gfla} (and other methods reported in \cite{gfla}) at 256$\times$256 resolution (our model is initially trained at 256$\times$176 and then fine-tuned to 256$\times$256). 
FID and LPIPS scores for methods with * are reproduced from GFLA, all other scores are computed by us using the same reference images used in \cite{gfla} (provided by the authors). Note that Intr-Flow is the only method leveraging 3D information. 
(b) Comparison with ADGAN \cite{adgan} at 256$\times$176 resolution. Arrows indicate whether higher ($\uparrow$) or lower ($\downarrow$) values of the metric are considered better.
}
\label{tab:pose_table}
\end{table}
\begin{figure}
    \centering
    \includegraphics[width=0.95\linewidth]{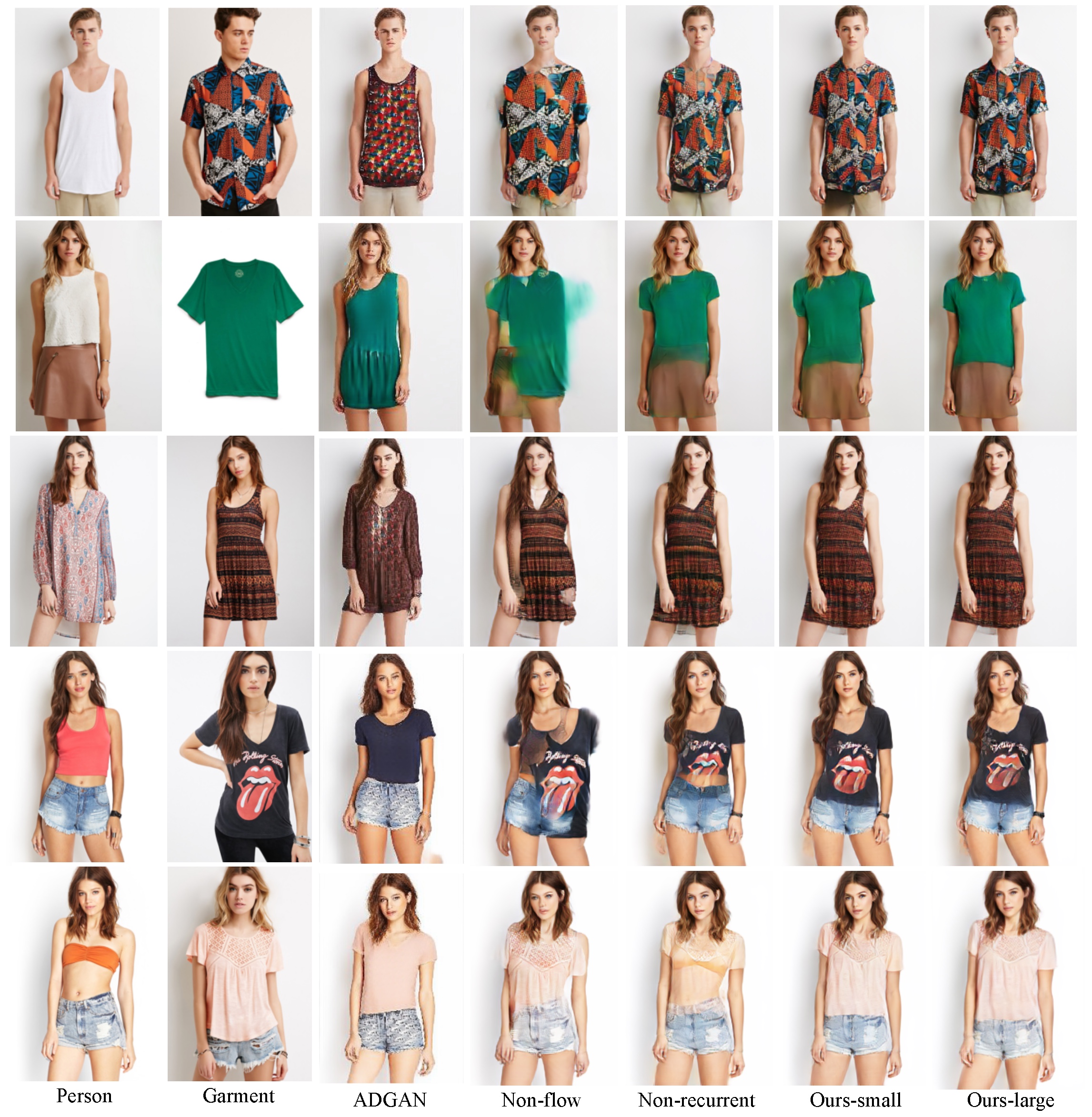}
    
    \caption{Virtual try-on results. We set the try-on order as (hair, bottom, top, jacket). Note that ADGAN is often unable to preserve the shape and texture of the transferred garment, while the non-recurrent version of our model creates ghosting in areas of garment overlap. In the second example, the garment is transferred from a ``shop'' image without a person, which is not our expected setting.}
    \label{fig:tryon}
\end{figure}

\subsection{Ablation Studies}

\noindent{\bf Recurrent generation.} 
We ablate our recurrent mechanism by merging the feature maps of all garments based on the softmax of their soft shape masks (without sigmoid taken) and injecting the merged feature map into a single-shot garment generator $\mathbf{G}_\mathrm{gar}$. As shown in Tab. \ref{tab:ablation}(a), the non-recurrent model gets a considerably lower sIoU than the full one. Fig. \ref{fig:tryon} shows the reason: when there is overlap between garments, the non-recurrent model tends to blend garments together, resulting in ghosting artifacts. 
\smallskip

\begin{figure}
    \centering
    \includegraphics[width=0.8\linewidth]{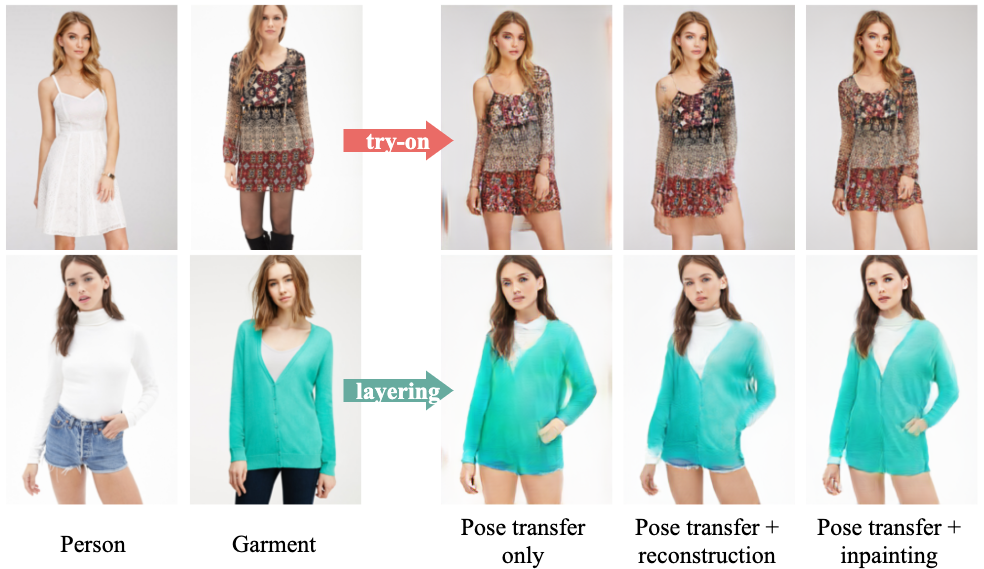}
    \caption{Comparison of our model trained on pose transfer only, joint with reconstruction, and joint with inpainting. The two ablated models have poorer ability to fill in holes (e.g., those created by hair on top of the source garment) or to create clear and coherent garment boundaries.}
    \label{fig:abl_joint}
    \vspace*{-15pt}
\end{figure}

\noindent{\bf Joint training on pose transfer and inpainting.} Tab. \ref{tab:ablation}(b) reports the results of our model trained without inpainting, and trained with reconstruction instead of inpainting (with $\alpha=0.2$). Although the differences from the full model are not apparent in the table for the pose transfer task, we can observe distinctive artifacts associated with different training choices in the rest applications like virtual try-on and layering, as described in section 3.3 and shown in Fig. \ref{fig:abl_joint}. 
\smallskip

\begin{figure}
    \centering
    \includegraphics[width=\linewidth]{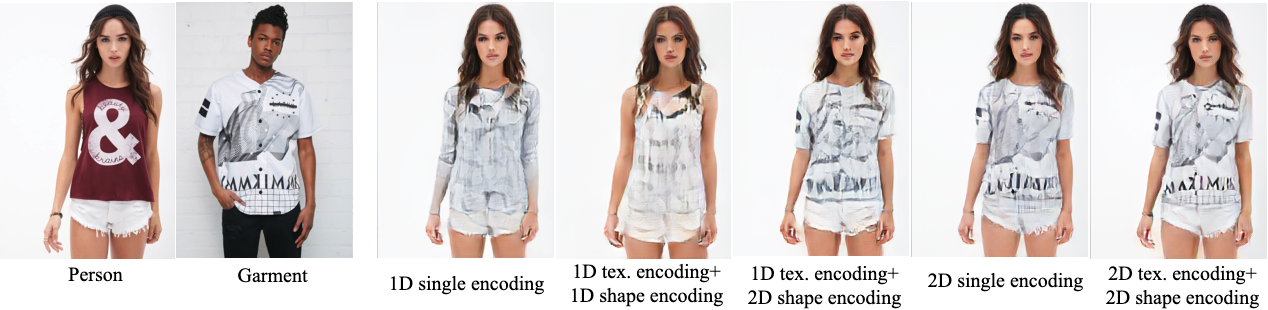} 
    \caption{Comparison of different garment encodings from Table \ref{tab:ablation}(c). The 1D encoding causes blurry texture. The single encoding in 2D is plausible in try-on but limits flexibility for editing tasks. }
    \vspace*{-15pt}
    \label{fig:joint_phi}
\end{figure}

\noindent{\bf Encoding: separate vs. single, 2D vs. 1D.}
We evaluate the effect of separate shape and texture encodings vs. a single garment encoding (i.e., joint shape and texture), as well as 2D vs. 1D encoding in Tab. \ref{tab:ablation}(c). To encode a garment by a single representation in 2D, we change Eq. (3) to $Z_k = \mathbf{\Phi}(Z_{k-1}, T_{g_k}) + Z_{k-1}$ to remove the shape factor. To further reduce this single encoding to 1D, we follow ADGAN's scheme (SPADE is 
switched to AdaIn~\cite{adain} in this case). For the version with separate 1D shape and texture codes, we attempt to decode the 1D shape vector into a mask by learning a segmenter consisting of a style block taking pose as input and generating the segmentation conditioned on the 1D shape vector, followed by broadcasting the texture vector into the shape mask to get the texture map. As an additional variant, we trained a model combining 1D texture encoding with 2D shape encoding, where we also get the texture map by broadcasting. 
From Tab. \ref{tab:ablation}(c) and Fig. \ref{fig:joint_phi}, we can see that the ablated versions, especially the 1D ones, are blurrier and worse at capturing details. This is consistent with our intuition that it is hard to recover spatial texture from a 1D vector. The 2D single encoding looks plausible but is still less sharp than the full model, plus it does not permit separate editing of shape and texture.

\begin{table}[]
\centering
\resizebox{0.95\columnwidth}{!}{
\begin{tabular}{|ll|l|l|l|l|}
\hline
&                    & SSIM$\uparrow$ & FID$\downarrow$ & LPIPS$\downarrow$ & sIoU$\uparrow$  \\ \hline \hline

& Full &   0.804 & 14.34 & \textbf{0.182} &  58.99  \\ \hline \hline

(a) & Non-recurrent &   0.804 & 14.85 & 0.183 &  58.44 \\ 
\hline
\hline
(b) & Pose transfer only training &   0.801 & \textbf{13.77} & 0.186 &  \textbf{59.01} \\ 
& Joint training with reconst. &   0.803 & 14.35 &  0.184 & 58.33  \\ \hline
\hline
(c) & Single 2D encoding &   \textbf{0.806} & 15.18 & 0.183 &  58.71 \\
& Single 1D encoding &  0.797  & 16.14 & 0.200 &  56.22 \\ 
& 1D texture+1D shape &  0.798  & 20.07 & 0.204 & 55.17  \\ 
& 1D texture+2D shape &   0.802 & 16.05 & 0.192 &  57.40 \\ \hline
(d) & Non-flow &   0.800 & 16.47 & 0.196 &  56.28 \\ 
\hline
\end{tabular}
}
\caption{
Ablation studies for the small model at 256$\times$176 on the pose transfer task (see text for details). 
}
\label{tab:ablation}
\end{table}

\noindent{\bf Flow field for garment localization.}
To prove the necessity of flow field $f$, which transforms the body part or garment from the source pose to its desired pose, we ablate the flow field $f$ by removing the global flow field estimator $F$ and the bilinear interpolation step in segment encoder $\mathbf{G}_\mathrm{seg}$. As evident from Fig. \ref{fig:tryon} and Tab. \ref{tab:ablation}(d), the flow field $f$ is essential for placing a garment in its right position and rendering the output realistically. 

\subsection{\bf User Study} 
Next, we report the results of a user study comparing our model to ADGAN and GFLA on pose transfer, and ADGAN on virtual try-on. We show users inputs and outputs from two unlabeled models in random order, and ask them to choose which output they prefer. 

For pose transfer, we randomly select 500 pairs from the test subset as the question pool. For virtual try-on, we randomly select 700 image pairs of person and garment (restricted to tops only), with the person facing front in both the person image and the garment image. We manually filter out the pairs which have no person shown in the person image, and which have no top shown in the garment image. For fairness, we also exclude all pairs with a jacket present, because our model considers jackets as separate garments from tops but ADGAN treats them as tops. When we run our model, the garment dressing order is set to (hair, top, bottom, jacket). Altogether, 22 questions for either pose transfer or try-on are given to each user. The first two questions are used as warm-up and not counted. We collected responses from 53 users for transfer, and 45 for try-on.

The results are shown in Table \ref{tab:user_study}. For pose transfer, our model is comparable to or slightly better than GFLA and ADGAN. Interestingly, ADGAN does not come off too badly on pose transfer despite its poor texture preservation because it tends to produce well-formed humans with few pose distortions, and generates nice shading and folds for textureless garments (see top example in Fig/ \ref{fig:pose}). For virtual try-on, the advantage of our model over ADGAN is decisive due to our superior ability to maintain the shape and texture of transferred garments (see Fig. \ref{fig:tryon}).
\begin{table}[]
\centering
\resizebox{0.8\columnwidth}{!}{%
\begin{tabular}{|l|l|l|}
\hline
  Compared method                & Task & Prefer other vs. ours \\ \hline \hline
GFLA \cite{gfla} &  pose transfer &  47.73\% vs. \textbf{52.27\%}  \\ 
ADGAN \cite{adgan} &  pose transfer &   42.52\% vs. \textbf{57.48\%}  \\ \hline
\hline
ADGAN \cite{adgan} &  virtual try-on & 19.36\% vs. \textbf{80.64\%}  \\ \hline
\end{tabular}
}
\caption{
User study results (see text). For fairness, ADGAN is compared with our large model trained at 256$\times$176, while GFLA is compared with our large model fine-tuned to 256$\times$256, with all outputs resized to 256$\times$176 before being displayed to users.
}
\label{tab:user_study}
\vspace*{-3pt}
\end{table}

\section{Editing Applications}
In this section, we demonstrate the usage of our model for several fashion editing tasks. With the exception of garment reshaping, which we found to need fine-tuning (see below), all the tasks can be done directly with the model trained as described in Section 3. See Supplemental Materials for additional qualitative examples.

\begin{figure}
    \centering
    \includegraphics[width=0.8\linewidth]{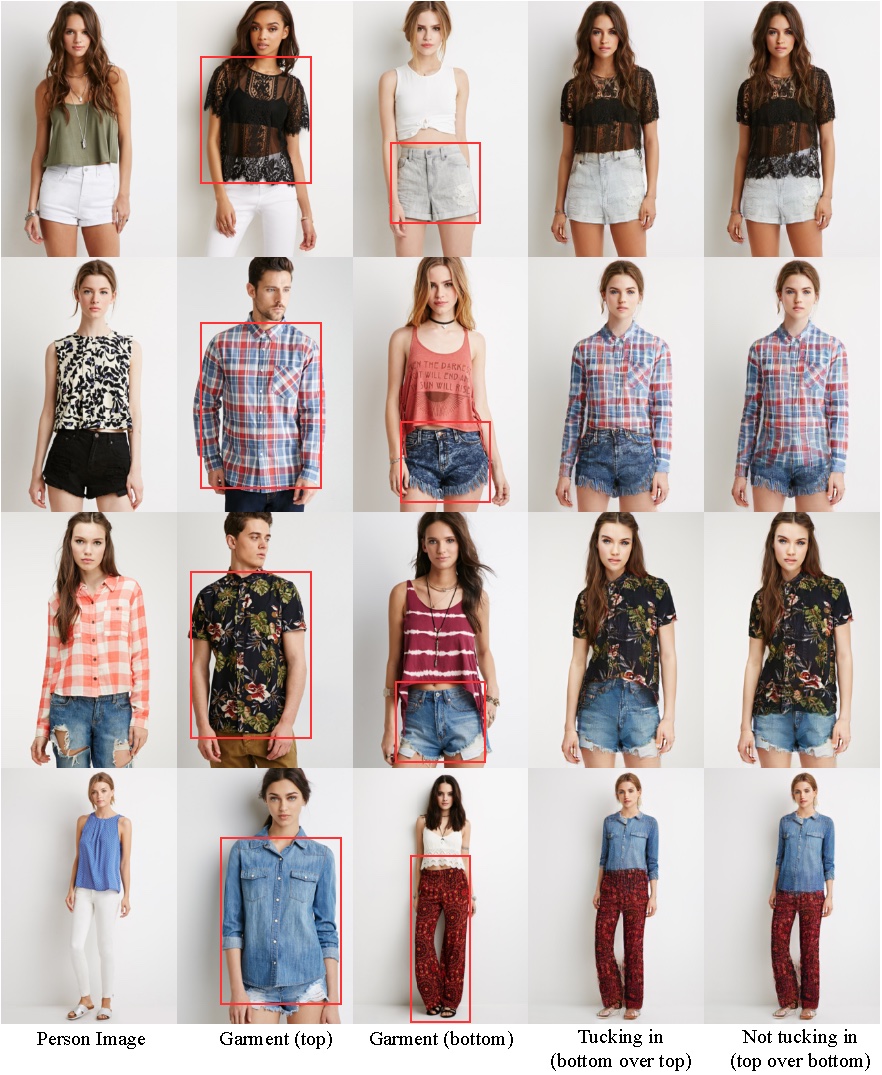}
    \vspace*{-5pt}
    \caption{{\bf Application: Tucking in.} Putting on the top {\em before} the bottom tucks it in, and putting it on {\em after} the bottom lets it out.}
    \label{fig:dress_in_order}
\end{figure}

\noindent{\bf Tucking in.}
As shown in Fig. \ref{fig:dress_in_order}, our model allows users to decide whether they want to tuck a top into a bottom by specifying dressing order.

\begin{figure}
    \centering
    \includegraphics[width=\linewidth]{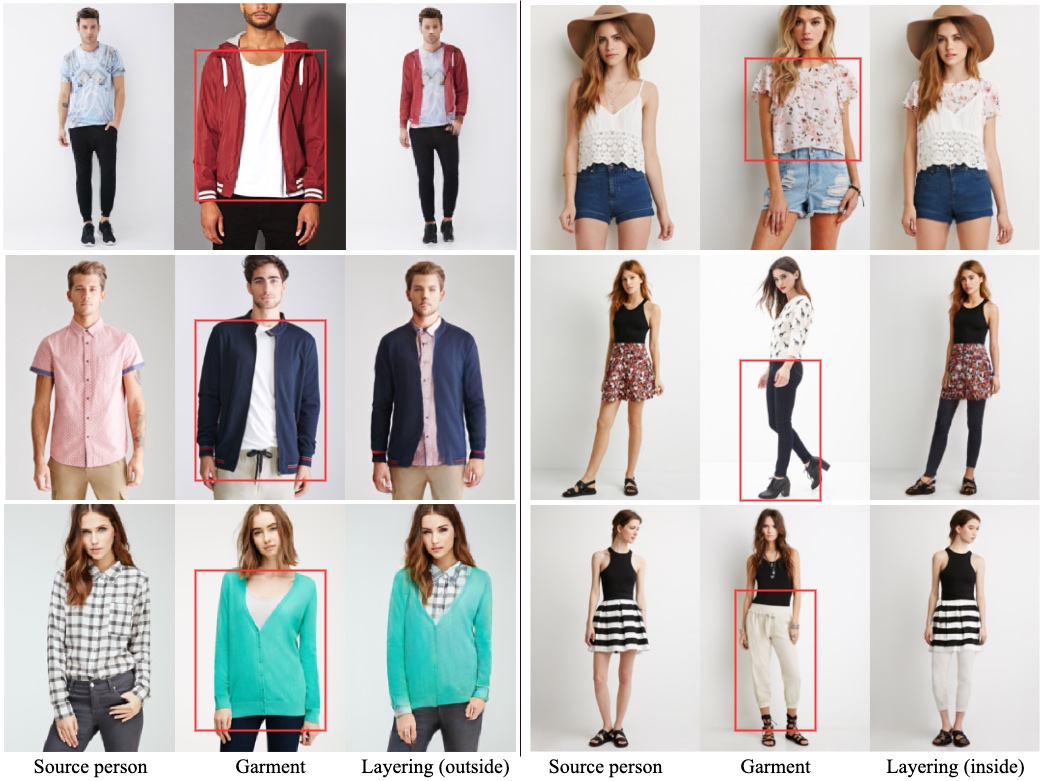}
    \caption{{\bf Application: Single layering.} Layer a garment outside (left) or inside (right) another garment in the same category.}
    \label{fig:dior_overlay1}
    \vspace*{-10pt}
\end{figure}

\begin{figure}
    \centering
    \includegraphics[width=0.8\linewidth]{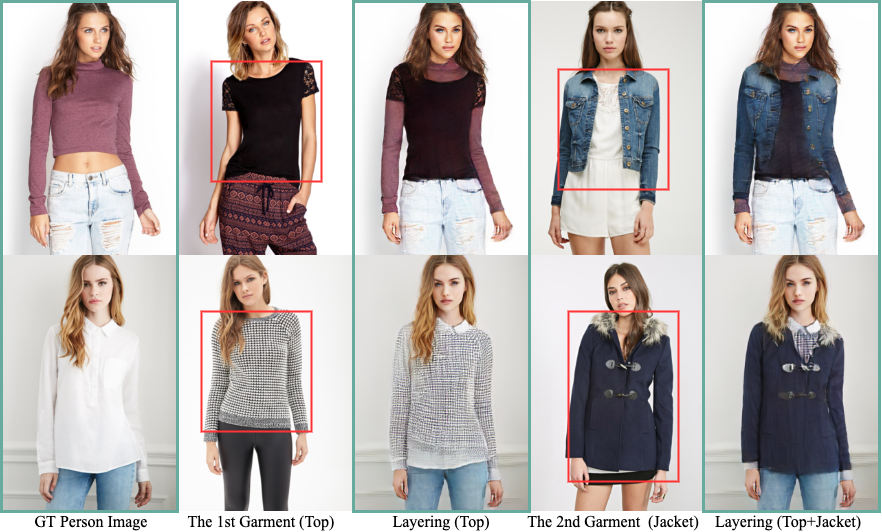}
    \vspace*{-1pt}
    \caption{{\bf Application: Double layering.} Layering two garments on top of the existing outfits in sequence. }
    \vspace*{-10pt}
    \label{fig:dior_overlay}
    
\end{figure}

\noindent{\bf Garment layering.} 
Fig. \ref{fig:dior_overlay1} shows the results of layering garments from the same category (top or bottom).
Fig. \ref{fig:dior_overlay} shows that we can also layer more than two garments in the same category (e.g., jacket over sweater over shirt).

\begin{figure}
    \centering
    \includegraphics[width=\linewidth]{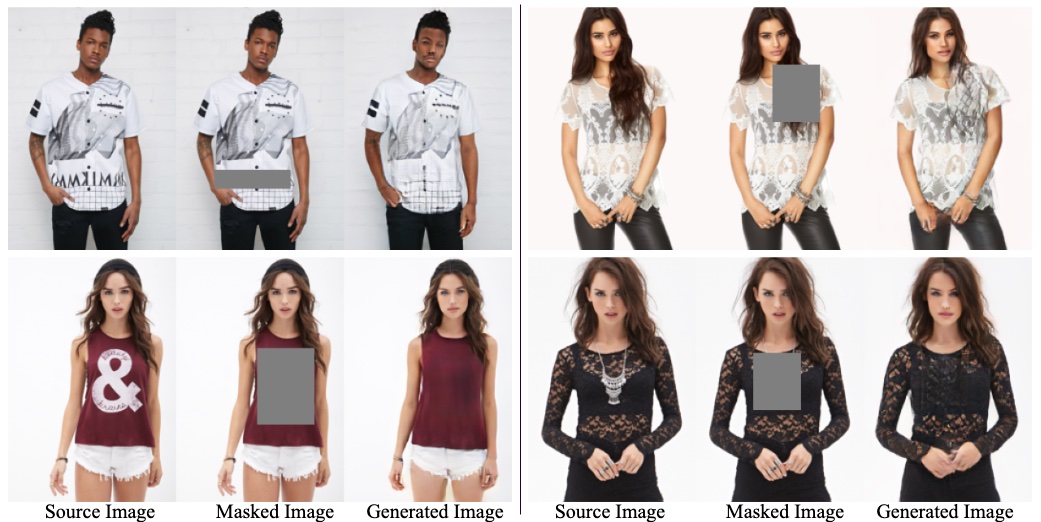}
    \caption{{\bf Application: Content removal.} }
    \vspace*{-10pt}
    \label{fig:app_removal}
\end{figure}

\noindent{\bf Content removal.}
To remove an unwanted print/pattern on a garment, we can mask the corresponding region in the texture map $T_{g}$ while keeping the shape mask $M_{g}$ unchanged, and the generator will fill in the missing part. (Fig. \ref{fig:app_removal}). 

\begin{figure}
    \centering
    \vspace*{-2pt}
    \includegraphics[width=\linewidth]{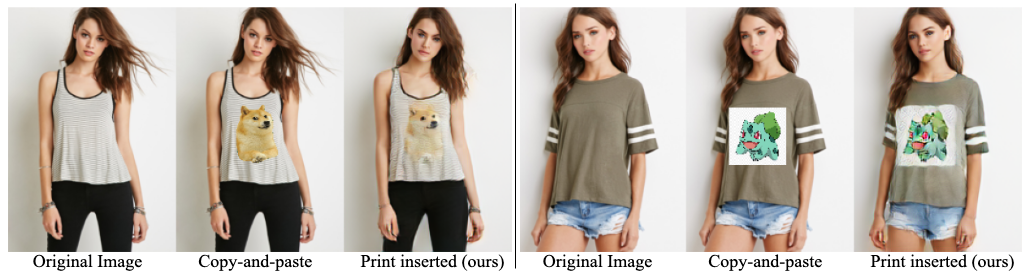}
    
    \caption{{\bf Application: Print insertion.}}
    \label{fig:app_insertion}
    \vspace*{-10pt}
\end{figure}

\noindent{\bf Print insertion.} 
To insert an external print, we treat the masked region from an external source as an additional ``garment''. 
In this case, the generation module is responsible for the blending and deformation, which limits the realism but produces plausible results as shown in Fig. \ref{fig:app_insertion}.

\begin{figure}
    \centering
    \includegraphics[width=\linewidth]{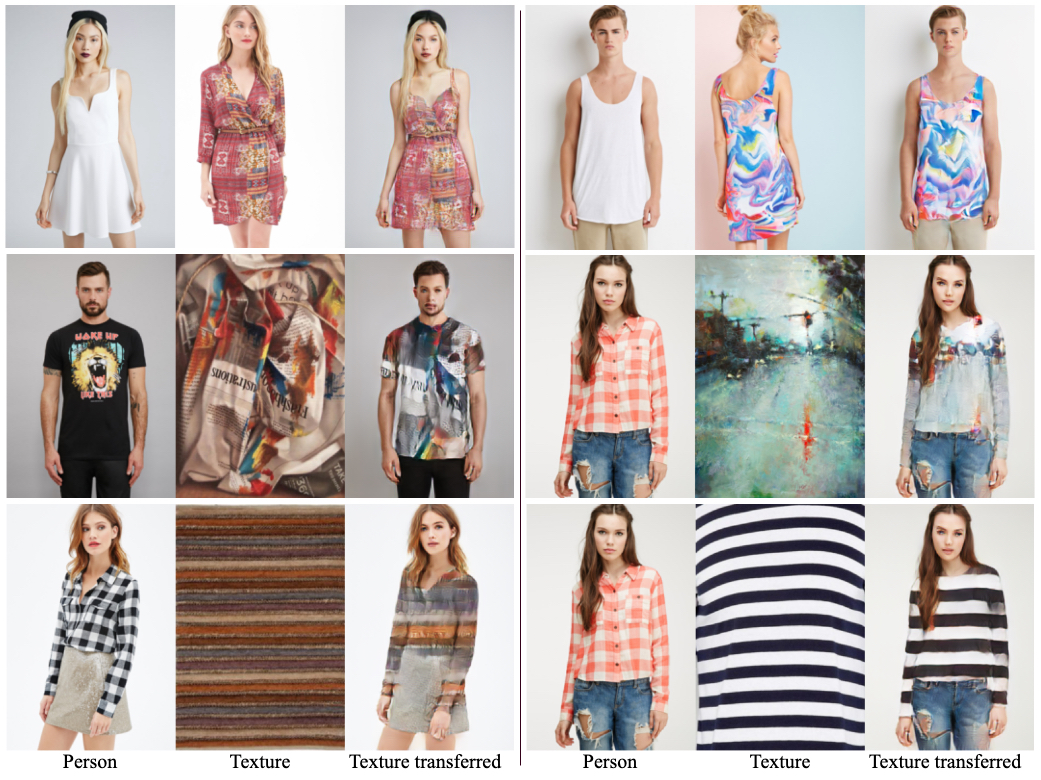}
    \vspace*{-2pt}
    \caption{{\bf Application: Texture transfer}. 
    }
    \vspace*{-10pt}
    \label{fig:app_tex_trans}
\end{figure}

\noindent{\bf Texture transfer.}
To transfer textures from other garments or external texture patches, we simply replace the garment texture map $T_g$ with the desired feature map encoded by $\mathbf{E}_\mathrm{tex}$. Fig.\ref{fig:app_tex_trans} shows the results of transferring textures from source garments (top row) and the DTD dataset \cite{cimpoi14describing} (bottom two rows). In the latter case, the texture does not deform over the body realistically, but the shading added by the generation module is plausible, and the results for less structured prints can be striking.

\begin{figure}
    \centering
    \includegraphics[width=\linewidth]{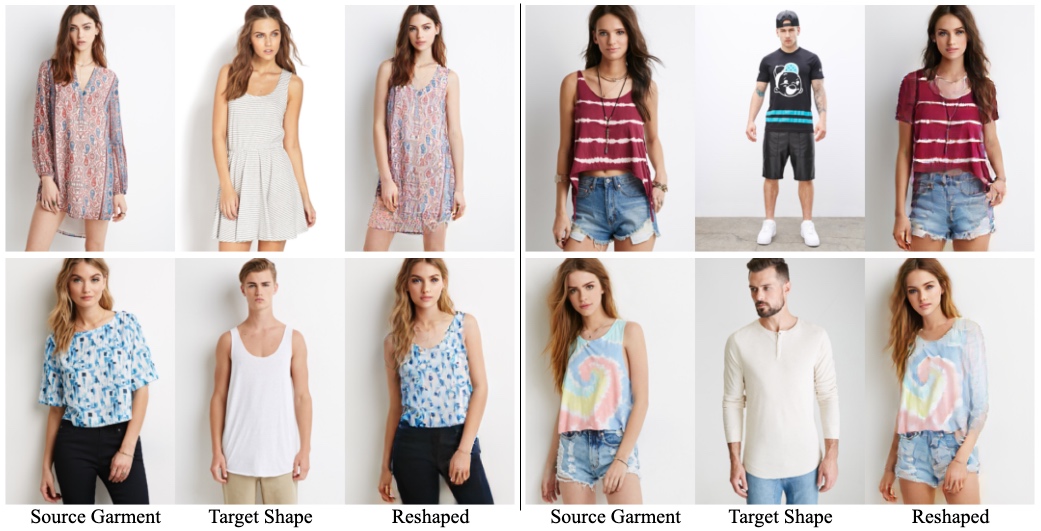}
    \caption{{\bf Application: Reshaping.}
    \vspace*{-10pt} }
    \label{fig:app_reshape}
\end{figure}

\noindent{\bf Reshaping.}
We can reshape a garment by replacing its shape mask with that of another garment (Fig. \ref{fig:app_reshape}). Our default model can easily handle removals (e.g. changing long sleeves to short), but not extensions (making sleeves longer). To overcome this, we fune-tuned the model with a larger inpainting ratio ($\alpha = 0.5$). The resulting model does a reasonable job of adding short sleeves to sleeveless garments (top right example of Fig. \ref{fig:app_reshape}), but is less confident in hallucinating long sleeves (bottom right).

\section{Limitations and Future Work}
\begin{figure}
    \centering
    \includegraphics[width=0.90\linewidth]{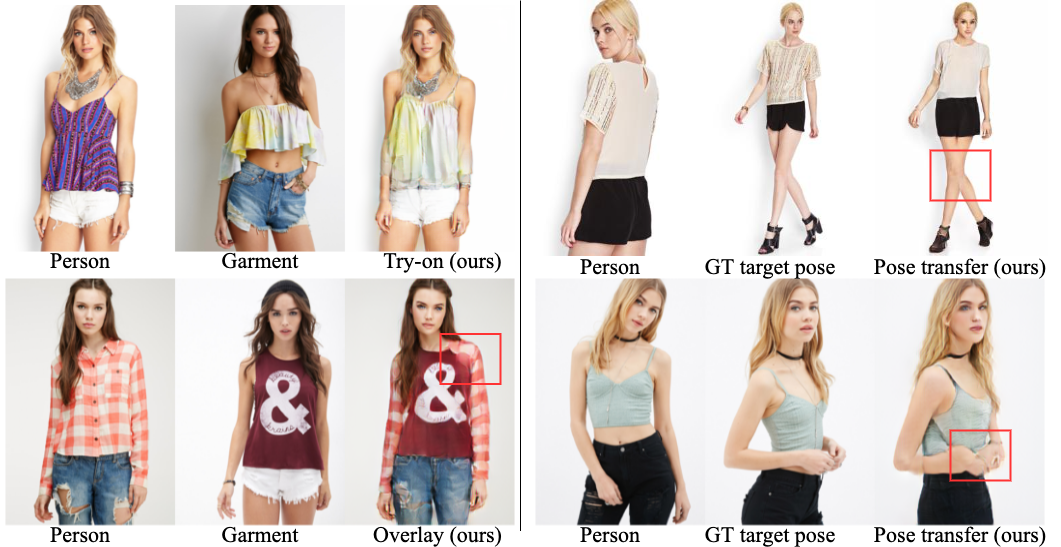}
    \caption{Failure cases. }
    \label{fig:failure}
    \vspace*{-2pt}
\end{figure}

This paper introduced DiOr, a flexible person generation pipeline trained on pose transfer and inpainting but capable of diverse garment layering and editing tasks for which there is no direct supervision. While our results are promising, there remain a number of limitations and failure modes. Some of these are illustrated in Fig.\ref{fig:failure}: complex or rarely seen poses are not always rendered correctly, unusual garment shapes are not preserved, some ghosting artifacts are present, and holes in garments are not always filled in properly. More generally, the shading, texture warping, and garment detail preservation of our method, while better than those of other recent methods, are still not entirely realistic. In the future, we plan to work on improving the quality of our output through more advanced warping and higher-resolution training and generation. \\

\noindent {\bf Acknowledgments.} This work was supported in part by NSF grants IIS 1563727
and IIS 1718221, Google Research Award, Amazon Research Award, and AWS Machine
Learning Research Award.

{\small
\bibliographystyle{ieee_fullname}
\bibliography{main}
}

\onecolumn
\appendix
\newpage
\section{User Study Interface}
Here we show the user interface from our user study. For both pose transfer and virtual try-on, 22 questions are presented to each user. Only one question is displayed at a time. When the users click the ``next question" button, they proceed to the next question and cannot go back. 
As shown in Figure \ref{supp_fig:ui}(a), for pose transfer, the users see a person in both source and target pose. Users are asked to choose the more realistic and accurate result from two generated images. The two options are randomly sorted, with one output coming from our large model and the other from one of the compared models. In Figure \ref{supp_fig:ui}(b), for virtual try-on, the users are provided with the reference person and target garment (upper-clothes). They are then once again asked to choose the better result from two randomly sorted generated images in terms of realism and accuracy.

\begin{minipage}{\linewidth}
\vspace{10pt}
\centering
\makebox[\linewidth]{
    \includegraphics[width=0.75\linewidth]{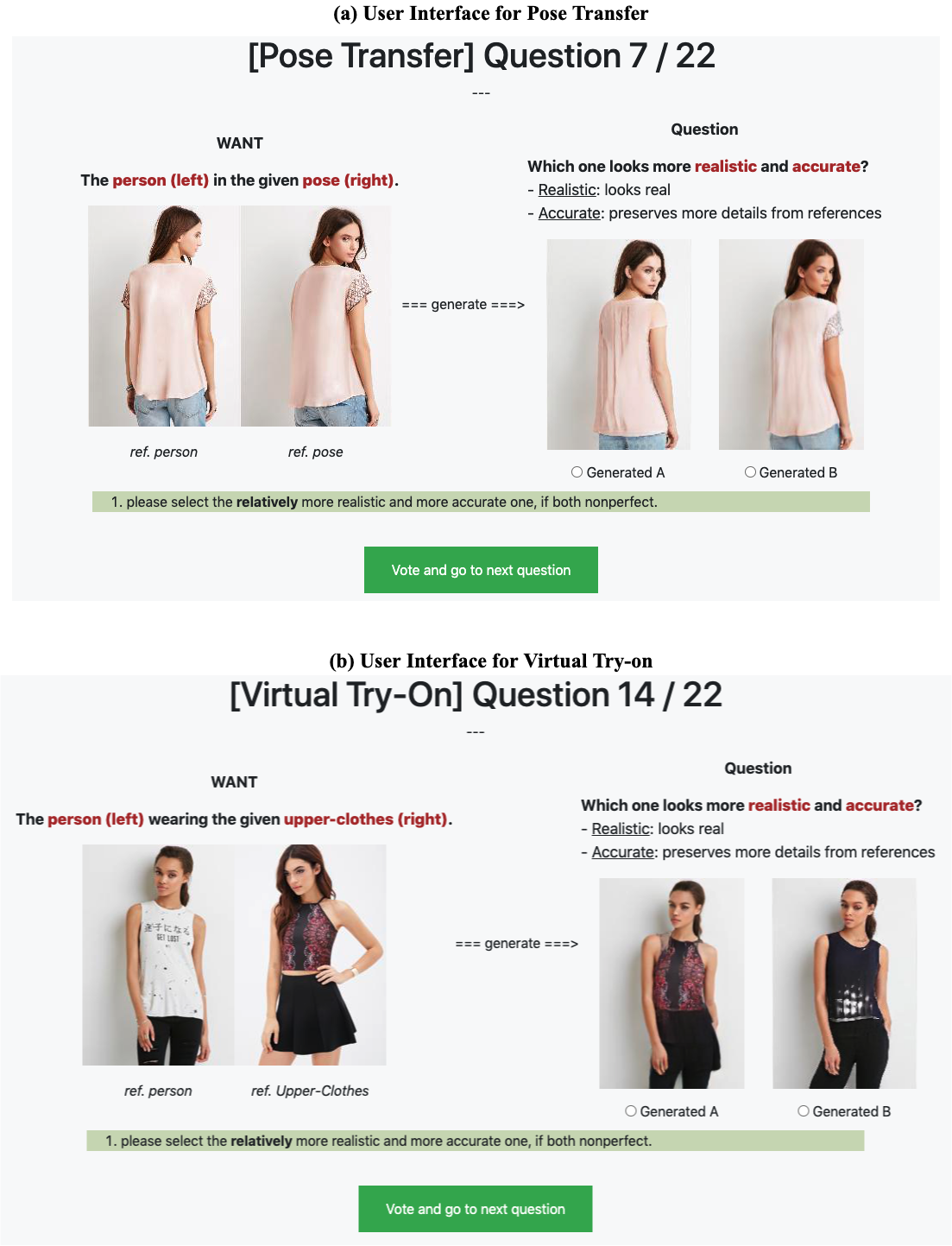}}
\captionof{figure}{ {\textbf{User Study Interface.} (a) User Interface for pose transfer. (b) User interface for virtual try-on.  }}
    \label{supp_fig:ui}
\end{minipage}

\newpage
\section{More Examples for Applications}
\label{appendix:example}
More examples are reported for each application as follows.

\subsection{Pose Transfer}
Figure \ref{supp_fig:pose} shows a random batch of pose transfer outputs from the test set. We include the ground truth, output of ADGAN \cite{adgan}, GFLA \cite{gfla}, our small model and our large model.

\begin{minipage}[c]{\linewidth}

    \makebox[\linewidth]{
        \includegraphics[width=0.7\linewidth]{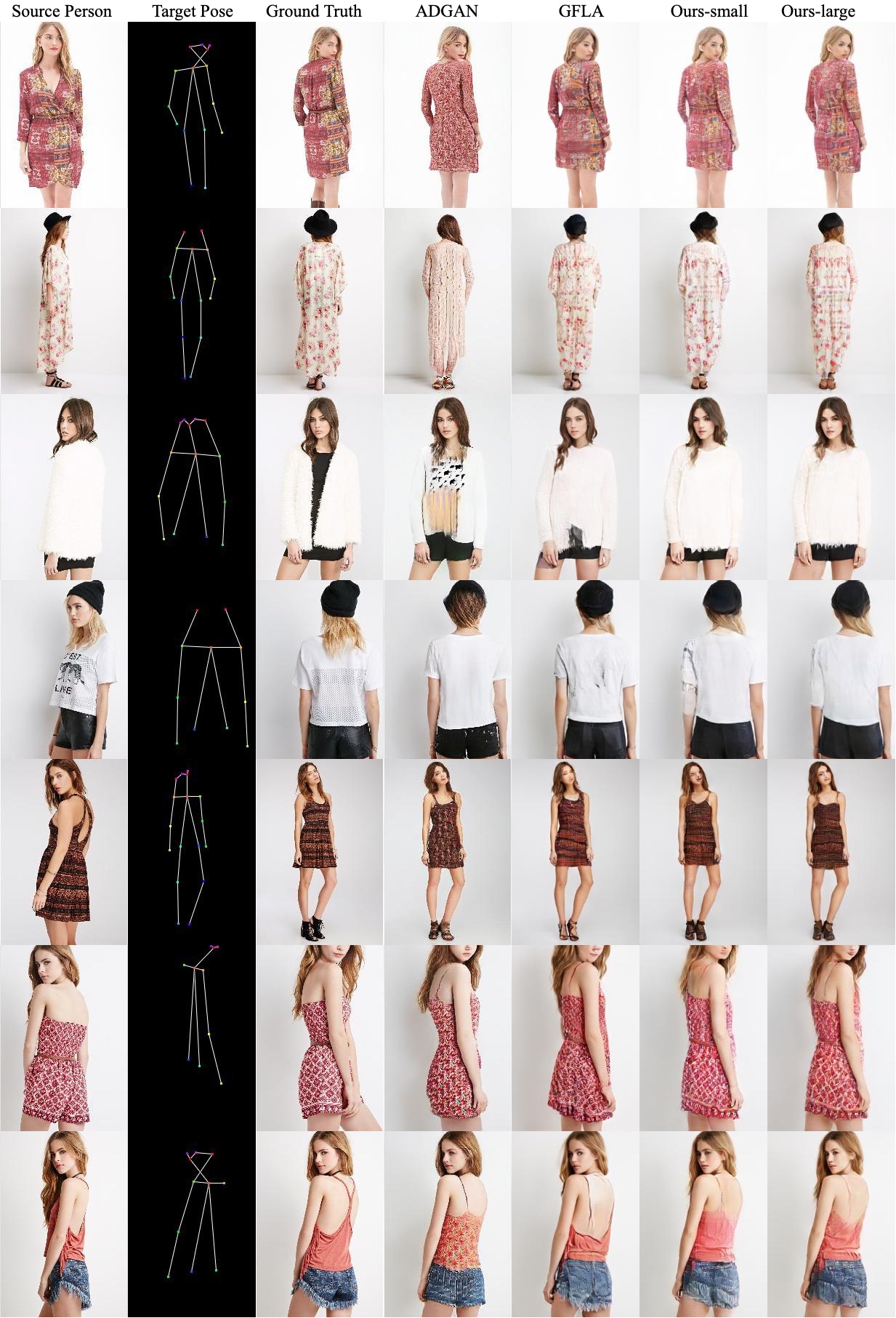}
    }
    \captionof{figure}{\textbf{Pose Transfer.}}
    \label{supp_fig:pose}
\end{minipage}

\newpage
\subsection{Virtual Try-on}
 For every person, we show try-on for two garments. As we have already shown the results of upper-clothes try-on, here we present dress try-on in Figure \ref{supp_fig:tryon}(a), pants try-on in Figure \ref{supp_fig:tryon}(b) and hair try-on in Figure \ref{supp_fig:tryon}(c). Note that we treat hair as a flexible component of a person and group it as a garment, so that we can freely change the hair style of a person. 
 
 In Figure \ref{supp_fig:tryon}, the first column is the target person, the next four columns include the first selected try-on garment with output results on the target person, and the last four columns include the second selected try-on garment with output results for the same target person. We provide generation results from ADGAN \cite{adgan}, our small model, and our large model.
 
 \begin{minipage}{\linewidth}
\centering
\makebox[\linewidth]{
  
    \includegraphics[width=0.9\linewidth]{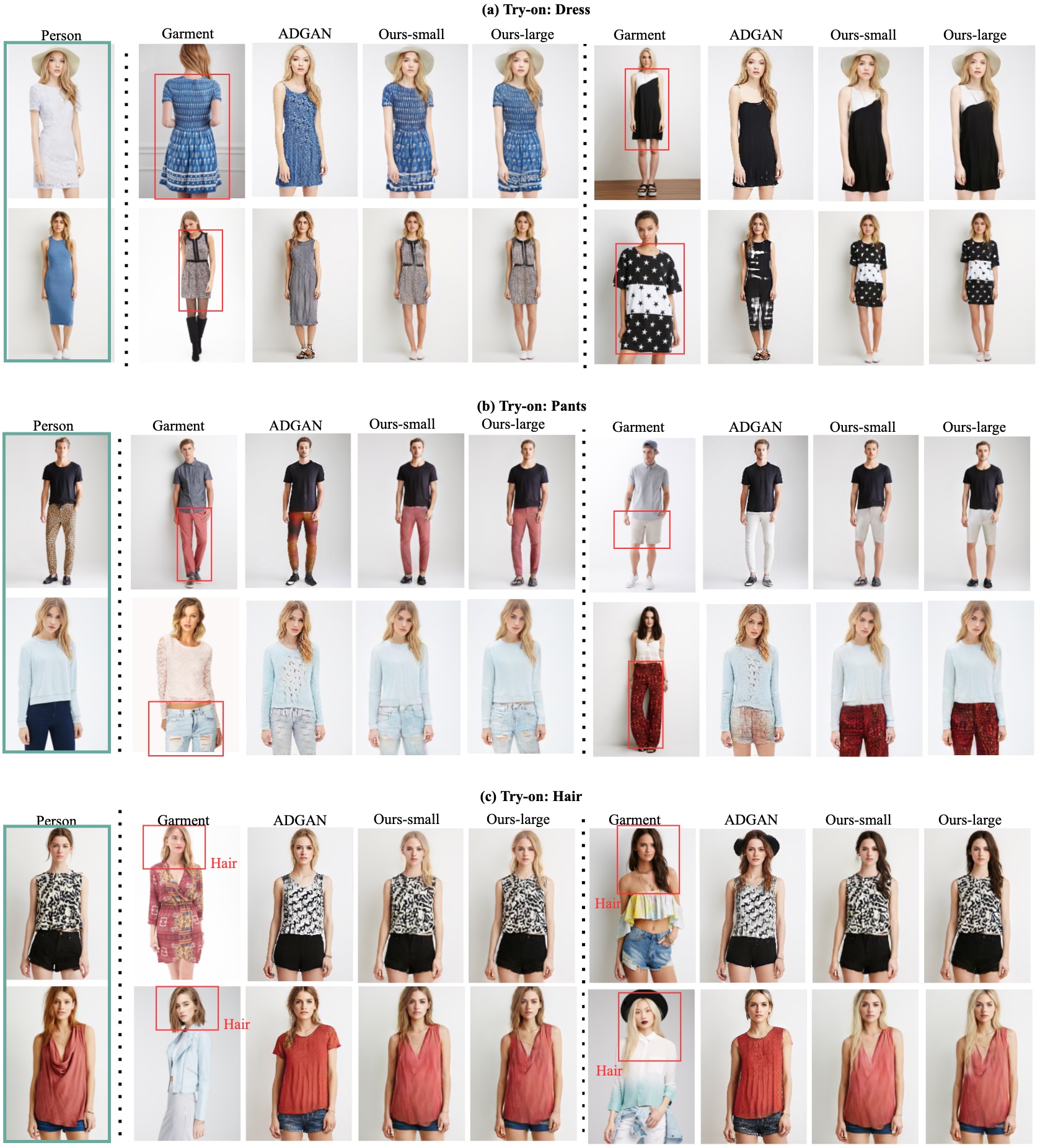}}
    \captionof{figure}{{\textbf{Virtual Try-on.}}}\label{supp_fig:tryon}
\end{minipage}

\newpage
\subsection{Dressing Order Effects}
We can achieve different looks from the same set of garments with different orders of dressing (e.g., tucking in or not). Figure \ref{supp_fig:tucking_in} demonstrates results from our large model for a person (first column) trying on a particular garment (the second column) with a different dressing order. Figure \ref{supp_fig:tucking_in}(a) shows the effect of dressing order for tucking or not, while Figure \ref{supp_fig:tucking_in}(b) demonstrates wearing a dress above or beneath a shirt.

 \begin{minipage}{\linewidth}
\centering
\makebox[\linewidth]{
  
    \includegraphics[width=\linewidth]{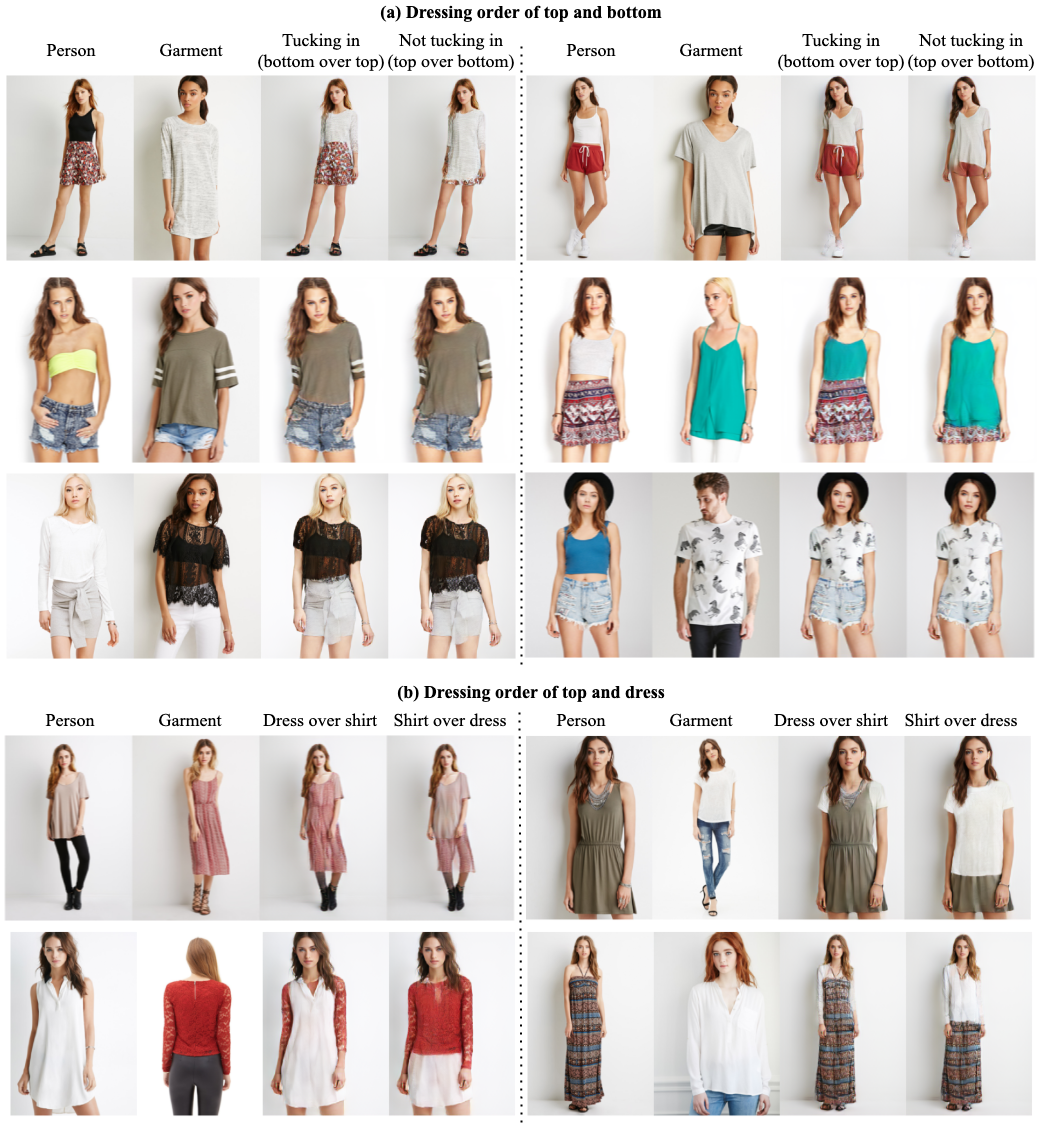}}
    \captionof{figure}{{\textbf{Dressing in order.}}}
    \label{supp_fig:tucking_in}
\end{minipage}

\newpage
\subsection{Layering}
 Here we include additional examples to demonstrate layering a single garment type. In Figure \ref{supp_fig:layering}(a), layering a new garment outside the existing garment is demonstrated on the left and layering a garment inside the existing garment is shown on the right. Figure \ref{supp_fig:layering} (b) shows more examples of layering two garments on top of the original garments.

\begin{minipage}{\linewidth}
\centering
\makebox[\linewidth]{
  
    \includegraphics[width=0.9\linewidth]{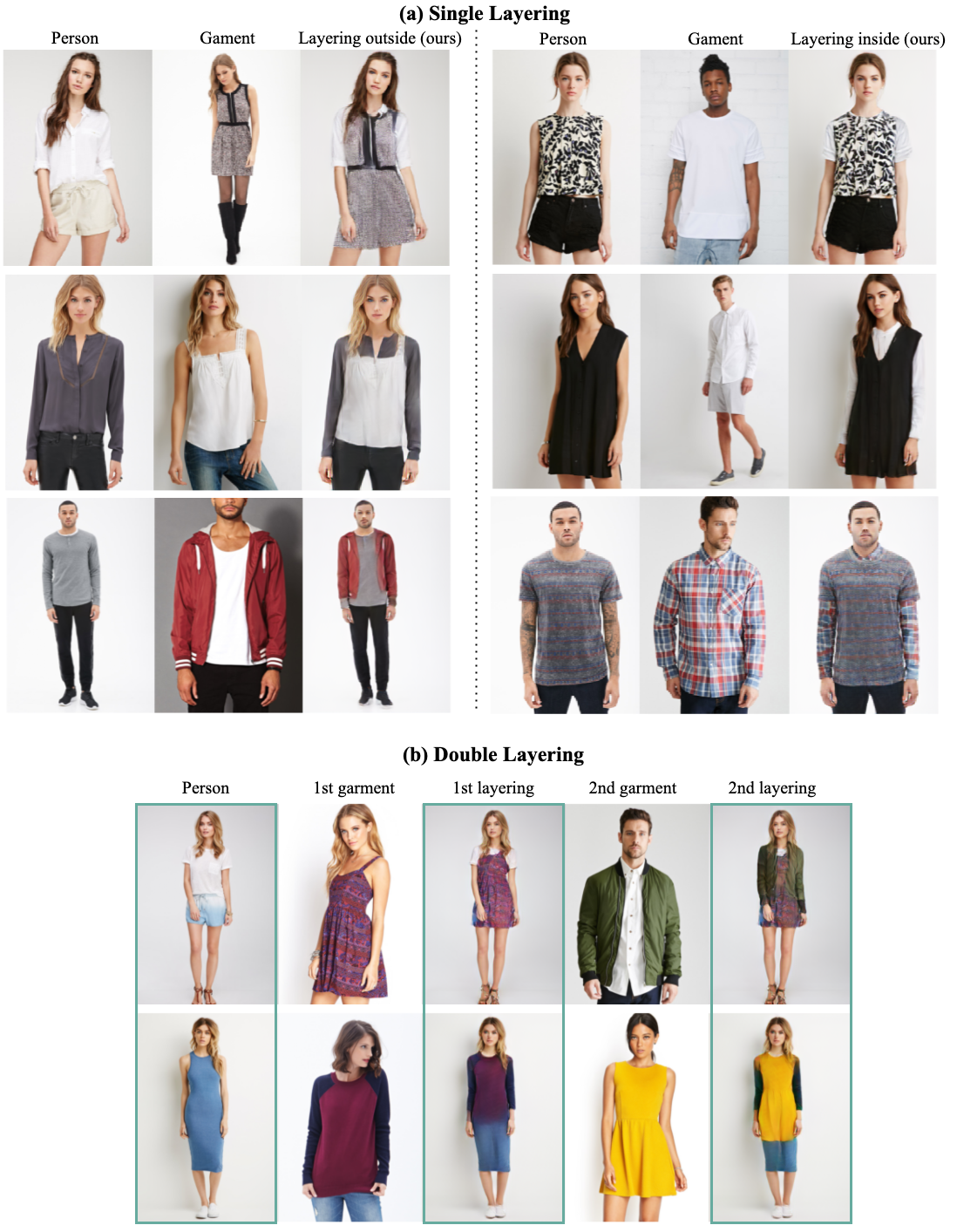}}
    \captionof{figure}{ {\textbf{Dressing in order.}}}
    \label{supp_fig:layering}
\end{minipage}

\newpage
\subsection{Content Removal}
To achieve content removal, we can mask out an unwanted region in the associated texture feature map for a garment. Results are shown in Figure \ref{supp_fig:removal}. In the bottom left example, although the girl's hair is partially masked out, we can remove only the pattern from the dress while keeping the hair, unlike the traditional inpainting methods. This is because the hair and dress are considered different garments and processed at different stages. This shows that our proposed person generation pipeline can better handle the relationships between garments.

\begin{minipage}{\linewidth}
\centering
\makebox[\linewidth]{
    \includegraphics[width=\linewidth]{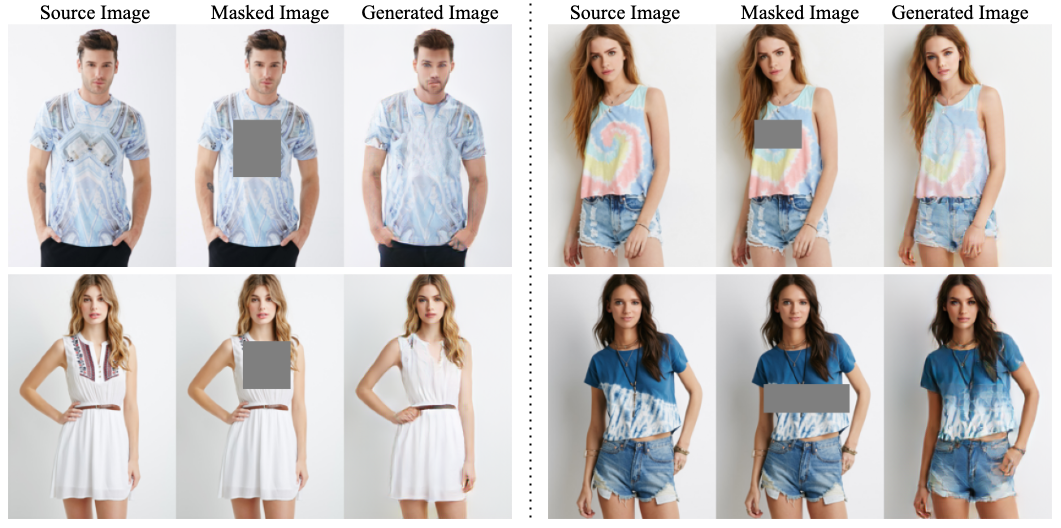}}
    \captionof{figure}{{\textbf{Content Removal}}}
    \label{supp_fig:removal}
\end{minipage}

\subsection{Print Insertion}
More results for print insertion are presented in Figure \ref{supp_fig:insertion}. From the left example, our model can warp a pattern onto existing garments, which is challenging for conventional harmonization methods. Additionally in the right example, although the print was placed partially on top of the hair, our novel pipeline can still render the hair in front of the inserted print.
This is done by setting our model processing order to first generate the print and then generate the hair.
\begin{minipage}{\linewidth}
\centering
\makebox[\linewidth]{
    \includegraphics[width=\linewidth]{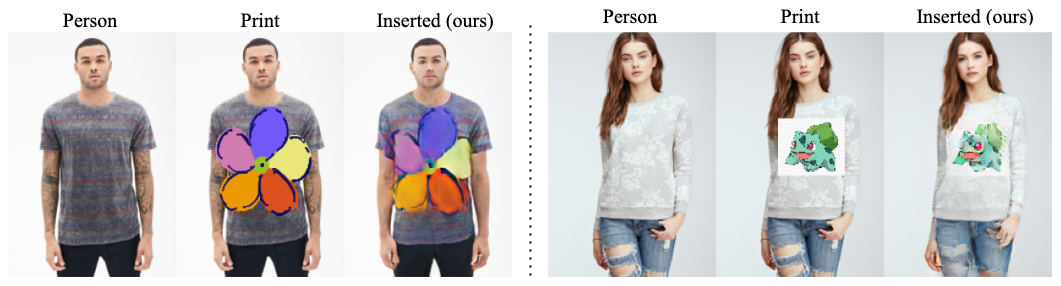}}
    \captionof{figure}{ {\textbf{Print Insertion.}}}
    \label{supp_fig:insertion}
\end{minipage}

\newpage
\subsection{Texture Transfer}
In Figure \ref{supp_fig:tex_trans}(a),  we can achieve texture transfer by switching the texture feature map $T_g$ for a garment. In Figure \ref{supp_fig:tex_trans}(b) we also transfer texture from external patches. We crop the texture from the patch using
    the soft mask $M_g$ (resized to image size) and encode the masked patch as the new texture feature map. We show results of naive crop and paste along with results produced by our large model.
    
\begin{minipage}{\linewidth}
\centering
\makebox[\linewidth]{
    \includegraphics[width=0.75\linewidth]{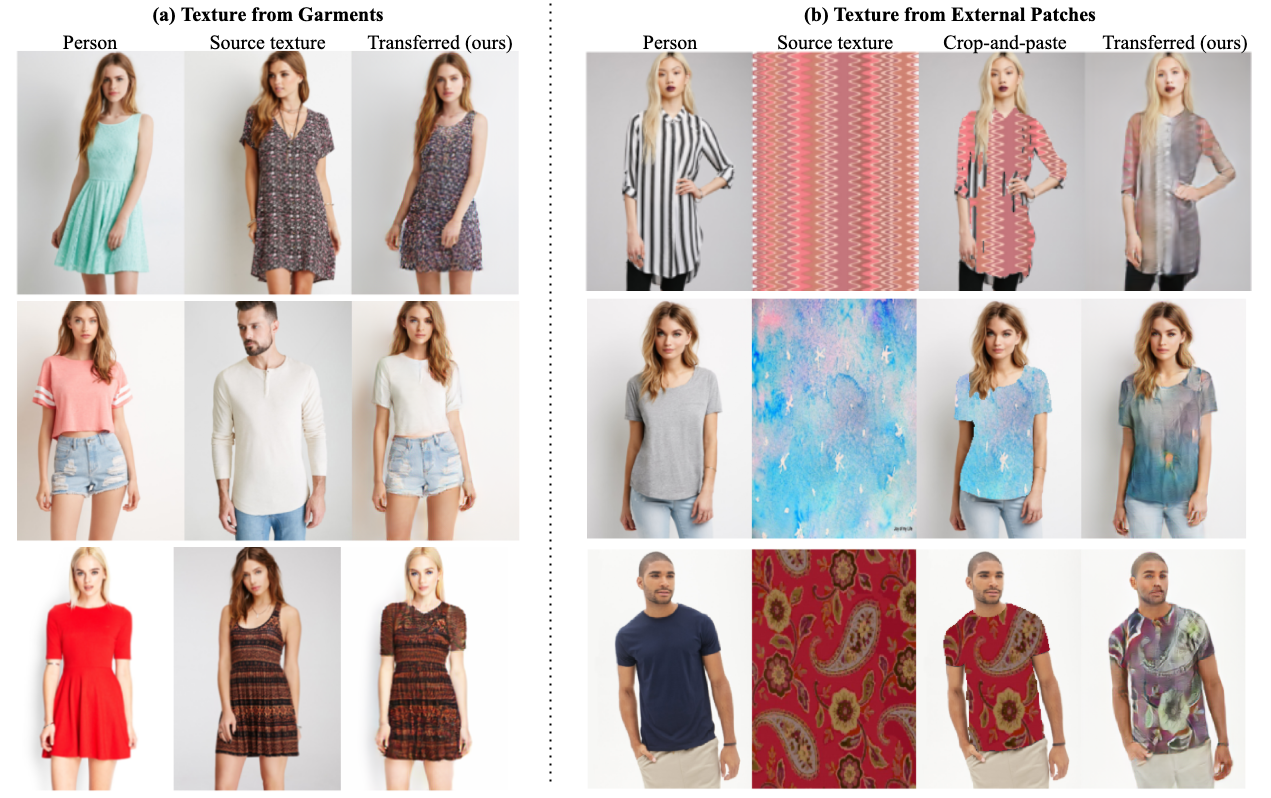}}
    \captionof{figure}{{\textbf{Texture Transfer.}}}
    \label{supp_fig:tex_trans}
\end{minipage}

\subsection{Reshaping}
We can reshape a garment by replacing its associated soft shape mask $M_g$ with the desired shape. In Figure \ref{supp_fig:reshaping}, the left column shows shortening of long garments, and the right column shows lengthening of short garments.

\begin{minipage}{\linewidth}
\centering
\makebox[\linewidth]{
    \includegraphics[width=0.7\linewidth]{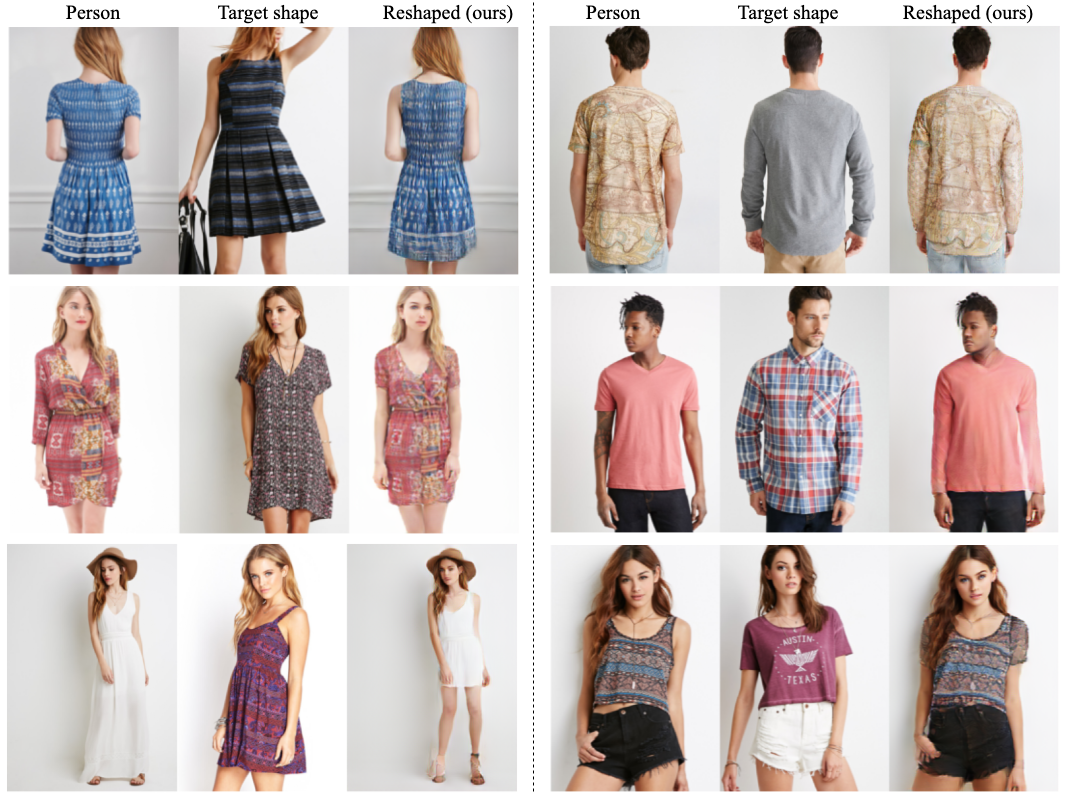}}
    \captionof{figure}{ {\textbf{Reshaping.}}}
    \label{supp_fig:reshaping}
\end{minipage}

\newpage
\section{Garment Transparency}
\label{appendix:transparency}
We also provide an investigation of how well the soft shape mask $M_g$ can control the transparency of garments in our DiOr system. In Figure \ref{supp_fig:transparency}, we show examples of a person trying on a new garment and then reapplying the original garment on top. When layering the original garment, we control its transparency by altering the soft shape mask $M_g$ with a transparency factor $a$ setting as $M_g[M_g > a] = a$.

For the first three examples, our method can control the transparency of the outermost garments well. However, in the fourth and the fifth examples, the lace parts on the garments become a solid peach color with increasing $a$. Ideally, these lace parts should show the color of the underlying garment in this case rather than the color of skin. 

\begin{minipage}{\linewidth}
\vspace{10pt}
\centering
\makebox[\linewidth]{
    \includegraphics[width=0.88\linewidth]{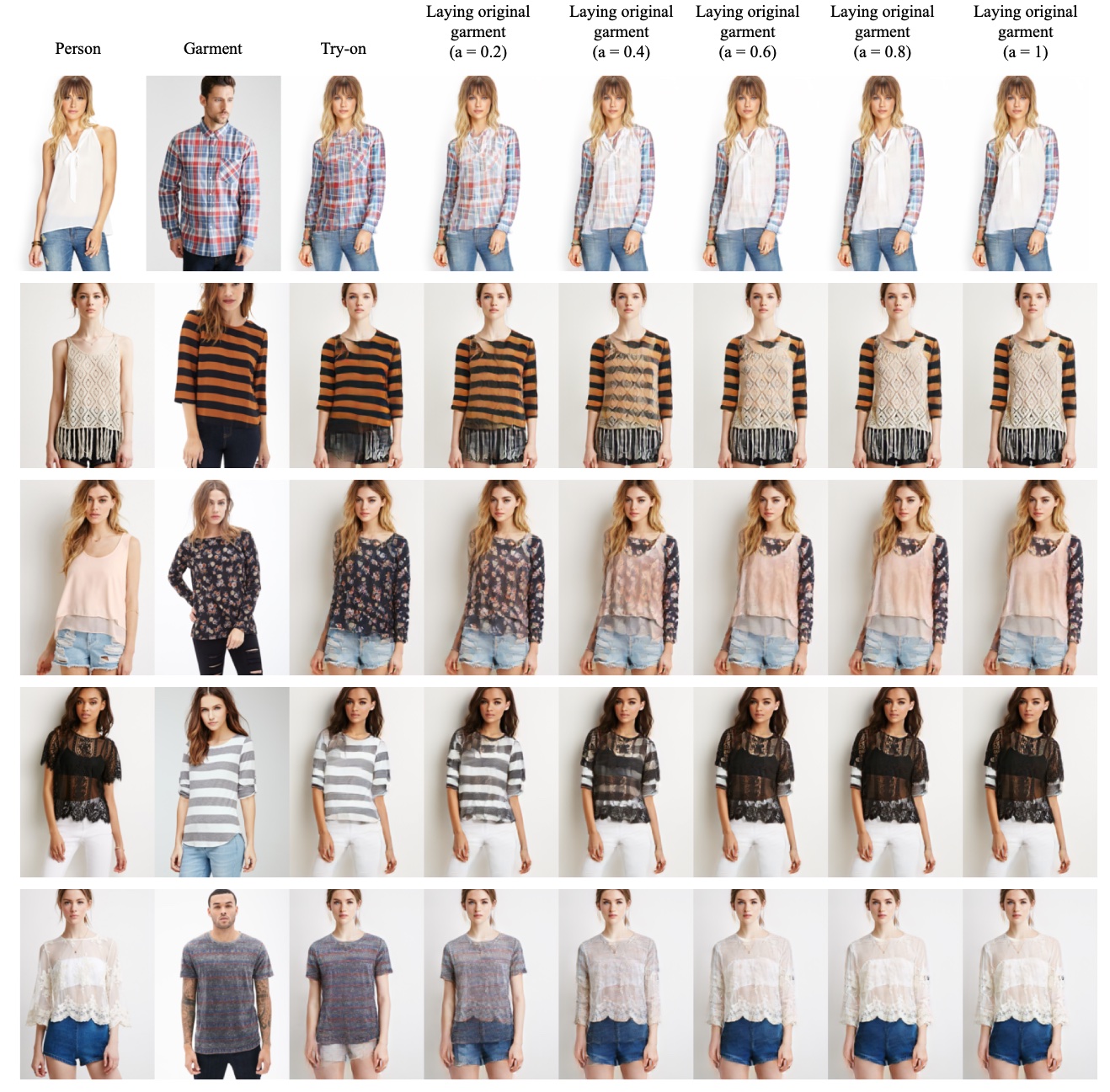}}
\captionof{figure}{ {\textbf{Transparency.} }}
    \label{supp_fig:transparency}
\end{minipage}

\end{document}